\documentclass{article}

\usepackage{arxiv}

\usepackage[utf8]{inputenc} 
\usepackage[T1]{fontenc}    
\usepackage{hyperref}       
\usepackage{url}            
\usepackage{amsfonts}       
\usepackage{nicefrac}       
\usepackage{graphicx}

\usepackage{multirow}
\usepackage{tabularx}
\usepackage{ragged2e}
\usepackage{enumitem}
\usepackage{booktabs} 
\usepackage{threeparttable} 
\usepackage{xtab}
\usepackage{longtable}
\usepackage{quoting}
\usepackage{subcaption}

\title{A Systematic Review of Aspect-based Sentiment Analysis: Domains, Methods, and Trends
\thanks {Published version in Artificial Intelligence Review: \url{http://dx.doi.org/10.1007/s10462-024-10906-z}}}

\date{26 July 2024}	

\author{
  Yan Cathy Hua \\
  School of Computer Science \\
  The University of Auckland \\ 
  New Zealand \\ 
  \texttt{yhua219@aucklanduni.ac.nz} \\
  \And
  Paul Denny \\
  School of Computer Science \\
  The University of Auckland \\ 
  New Zealand \\ 
  \texttt{p.denny@auckland.ac.nz} \\
  \And
  Katerina Taskova \\
  School of Computer Science \\
  The University of Auckland \\ 
  New Zealand \\ 
  \texttt{katerina.taskova@auckland.ac.nz} \\
  \And
  J{\"o}rg Wicker \\
  School of Computer Science \\
  The University of Auckland \\ 
  New Zealand \\ 
  \texttt{j.wicker@auckland.ac.nz} \\
}

\renewcommand{\headeright}{}
\renewcommand{\undertitle}{}
\renewcommand{\shorttitle}{A Systematic Review of Aspect-based Sentiment Analysis (ABSA)}

\hypersetup{
pdftitle={A Systematic Review of Aspect-based Sentiment Analysis (ABSA)},
pdfauthor={Yan (Cathy) ~Hua, Paul ~Denny, Katerina ~Taskova, J{\"o}rg ~Wicker},
pdfkeywords={ABSA, aspect-based sentiment analysis, systematic literature review, natural language processing},
}

\begin{document}
\maketitle

\begin{abstract} 
Aspect-based Sentiment Analysis (ABSA) is a fine-grained type of sentiment analysis that identifies aspects and their associated opinions from a given text. With the surge of digital opinionated text data, ABSA gained increasing popularity for its ability to mine more detailed and targeted insights. Many review papers on ABSA subtasks and solution methodologies exist, however, few focus on trends over time or systemic issues relating to research application domains, datasets, and solution approaches. To fill the gap, this paper presents a Systematic Literature Review (SLR) of ABSA studies with a focus on trends and high-level relationships among these fundamental components. This review is one of the largest SLRs on ABSA. To our knowledge, it is also the first to systematically examine the interrelations among ABSA research and data distribution across domains, as well as trends in solution paradigms and approaches. Our sample includes 727 primary studies screened from 8550 search results without time constraints via an innovative automatic filtering process. Our quantitative analysis not only identifies trends in nearly two decades of ABSA research development but also unveils a systemic lack of dataset and domain diversity as well as domain mismatch that may hinder the development of future ABSA research. We discuss these findings and their implications and propose suggestions for future research.
\end{abstract}



\section{Introduction}\label{sec1_intro}

In the digital era,  a vast amount of online opinionated text is generated daily through which people express views and feelings (i.e. sentiment) towards certain subjects, such as user reviews, social media posts, and open-ended survey question responses  \cite{ref1}. Understanding the sentiment of these opinionated text data is essential for gaining insights into people's preferences and behaviours and supporting decision-making across a wide variety of domains  \cite{ref2,OR15,8,55,72}. The analyses of opinionated text usually aim at answering questions such as ``\textit{What subjects were mentioned?}'', ``\textit{What did people think of (a specific subject)?}'', and ``\textit{How are the subjects and/or opinions distributed across the sample?}'' (e.g.  \cite{252,ref11,ref12,ref13}). These objectives, along with today’s enormous volume of digital opinionated text, require an automated solution for identifying, extracting and classifying the subjects and their associated opinions from the raw text. Aspect-based Sentiment Analysis (ABSA) is one such solution. 

\subsection{Review Focus and Research Questions}

This work presents a Systematic Literature Review (SLR) of existing ABSA studies with a large-scale sample and quantitative results. We focus on trends and high-level patterns instead of methodological details that were well covered by the existing surveys mentioned above. We aim to benefit both ABSA newcomers by introducing the basics of the topic, as well as existing ABSA researchers by sharing perspectives and findings that are useful to the ABSA community and can only be obtained beyond the immediate research tasks and technicalities.

We seek to answer the following sets of Research Questions (RQs):
\begin{itemize}
    \item[] \textbf{RQ1.} To what extent is ABSA research and its dataset resources dominated by the commercial (especially the product and service review) domain? What proportion of ABSA research focuses on other domains and dataset resources?
    
    \item[] \textbf{RQ2.} What are the most common ABSA problem formulations via subtask combinations, and what proportion of ABSA studies only focus on a specific subtask?
    
    \item[] \textbf{RQ3.} What is the trend in the ABSA solution approaches over time? Are linguistic and traditional machine-learning approaches still in use?
\end{itemize}

\vspace{\baselineskip}

This review makes a number of unique contributions to the ABSA research field: 1) It is one of the largest scoped SLRs on ABSA,
with a main review and a Phase-2 targeted review of a combined 727 primary studies published in 2008-2024, selected from 8550 search results without time constraint. 2) To our knowledge, it is the first SLR that systematically examines the ABSA data resource distribution in relation to research application domains and methodologies; and 3) Our review methodology adopted an innovative automatic filtering process based on PDF-mining, which enhanced screening quality and reliability. Our quantitative results not only revealed trends in nearly two decades of ABSA research literature but also highlighted potential systemic issues that could limit the development of future ABSA research.

\subsection{Organisation of This Review}

In Section \ref{sec2_background} (``Background''), we introduce ABSA and highlight the motivation and uniqueness of this review. Section \ref{sec3_methods} (``Methods'') outlines our SLR procedures, and Section \ref{sec4_results} (``Results'') answers the research questions with the SLR results. We then discuss the key findings and acknowledge limitations in Sections \ref{sec5_discussion} and \ref{sec6_conclusion} (``Discussion'' and ``Conclusion''). 

For those interested in more details, Appendix \ref{appendix_A_absa} provides an in-depth introduction to ABSA and its subtasks. Appendix \ref{appendix_B_method} describes the full details of our Methods, and additional figures from the Results are provided in Appendix\ref{appendix_C_results}. 

\section{Background} \label{sec2_background}

\subsection{ABSA: a fine-grained sentiment analysis} \label{2.1_absa}

Aspect-based Sentiment Analysis (ABSA) is a sub-field of Sentiment Analysis (SA), which is a core task of natural language processing (NLP). SA, also known as ``opinion mining''  \cite{55,72,150,167,237}, solves the problem of identifying and classifying given text corpora’s affect or sentiment orientation  \cite{2,8} into polarity categories (e.g. ``positive, neutral, negative'')  \cite{OR3,OR4,OR5}, intensity/strength scores (e.g. from 1 to 5)  \cite{387}, or other categories. The ``\textit{identifying the subjects of opinions}'' part of the quest relates to the granularity of SA. Traditional SA mostly focuses on document- or sentence-level sentiment and thus assumes a single subject of opinions  \cite{RV9,RV11}. In recent decades, the explosion of online opinion text has attracted increasing interest in distilling more targeted insights on specific entities or their aspects within each sentence through finer-grained SA  \cite{RV9,RV11,2,127,159}. This is the problem ABSA aims to solve. 

\subsection{ABSA and its subtasks} \label{2.2_subtasks}

ABSA involves identifying the sentiments toward specific entities or their attributes, called \textbf{aspects}. These aspects can be explicitly mentioned in the text or implied from the context (``implicit aspects''), and can be grouped into aspect categories \cite{1,2,RV1,374,174,195}. \textbf{Appendix \ref{A1_absa}} presents a more detailed definition of ABSA, including its key components and examples.  

A complete ABSA solution as described above traditionally involves a combination of subtasks, with the fundamental ones  \cite{13,19,70,253, 207} being Aspect (term) Extraction (AE), Opinion (term) Extraction (OE), and Aspect-Sentiment Classification (ASC), or in an aggregated form via Aspect-Category Detection
(ACD) and Aspect Category Sentiment Analysis (ACSA). 

The choice of subtasks in an ABSA solution reflects both the problem formulation and, to a large extent, the technologies and resources available at the time. The solutions to these fundamental ABSA subtasks evolved from pure linguistic and statistical solutions to the dominant machine learning (ML) approaches \cite{RV1,RV5,RV11,330}, usually with multiple subtask models or modules orchestrated in a pipeline \cite{21,71}. More recently, the rise of multi-task learning brought an increase in End-to-end (E2E) ABSA solutions that can better capture the inter-task relations via shared learning \cite{batch2_01}, and many only involve a single model that provides the full ABSA solution via one composite task \cite{19,21,OR1}. The most typical composite ABSA tasks include Aspect-Opinion Pair Extraction (AOPE) \cite{71,87,139}, Aspect-Polarity Co-Extraction (APCE) \cite{19,OR6}, Aspect-Sentiment Triplet Extraction (ASTE) \cite{19,21,152,253}, and Aspect-Sentiment Quadruplet Extraction/Prediction (ASQE/ASQP) \cite{74,372,OR25,ref-PARAPHRASE}. We provide a more detailed introduction to ABSA subtasks in \textbf{Appendix \ref{A2_subtasks}}.

\subsection{The context- and domain-dependency challenges} \label{2.3_domainDependency}

The nature and the interconnection of its components and subtasks determine that ABSA is heavily domain- and context-dependent  \cite{RV9,RV99,111}. Domain refers to the ABSA task (training or application) topic domains, and context can be either the ``global'' context of the document or the ``local'' context from the text surrounding a target word token or word chunks. At least in English, the same word or phrase could mean different things or bear different sentiments depending on the context and topic domains. For example, ``a big fan'' could be an electric appliance or a person, depending on the sentence and the domain; ``cold'' could be positive for ice cream but negative for customer service; and ``DPS'' (damage per second) could be either a gaming aspect or non-aspect in other domains. Thus, the ability to incorporate relevant context is essential for ABSA solutions; and those with zero or very small context windows, such as n-gram and Markov models, are rare in ABSA literature and can only tackle a limited range of subtasks (e.g.  \cite{478}). 

Moreover, although many language models (e.g. Bidirectional Encoder Representations from Transformers (BERT,  \cite{refBERT}), Generative pre-trained transformers (GPT,  \cite{refGPT3}), Recurrent Neural Network (RNN)-based models) already incorporated local context from the input-sequence and/or general context through pre-trained embeddings, they still performed unsatisfactorily on some ABSA domains and subtasks, especially Implicit AE (IAE), AE with multi-word aspects, AE and ACD on mixed-domain corpora, and context-dependent ASC  \cite{ref3,127,150,111}. Many studies showed that ABSA task performance benefits from expanding the feature space beyond the generic and input textual context. This includes incorporating domain-specific dataset/representations and additional input features such as Part-of-Speech (POS) tags, syntactic dependency relations, lexical databases, and domain knowledge graphs or ontologies  \cite{111,127,150}. Nonetheless, annotated datasets and domain-specific resources are costly to produce and limited in availability, and domain adaptation, as one solution to this, has been an ongoing challenge for ABSA  \cite{17,OR1,RV9,111,batch2_187}.

The above highlights the critical role of domain-specific datasets and resources in ABSA solution quality, especially for supervised approaches. On the other hand, it suggests the possibility that the prevalence of dataset-reliant solutions in the field, and a skewed ABSA dataset domain distribution, could systemically hinder ABSA solution performance and generalisability  \cite{17,6}, thus confining ABSA research and solutions close to the resource-rich domains and languages. This idea underpins this literature review’s motivation and research questions.

\subsection{Review Rationale} \label{2.4_rationale}

This review is motivated by the following rationales:

First, the shift towards ML, especially supervised and/or DL solutions for ABSA, highlights the importance of dataset resources. In particular, annotated large benchmark datasets are crucial for the quality and development of ABSA research. Meanwhile, the finer granularity of ABSA also brings the persistent challenge of domain dependency described in Subsection \ref{2.3_domainDependency}. The diversity of ABSA datasets and their domains can have a direct and systematic impact on research and applications. 

The early seminal works in ABSA were motivated by commercial applications and focused on product and service reviews  \cite{RV11,RV21,RV19}, such as Ganu et al., (2009)  \cite{Ganu2009}, Hu and Liu (2004a  \cite{HuLiu2004a}, 2004b  \cite{HuLiu2004b}), and Pontiki et al. (2014  \cite{semeval2014}, 2015  \cite{semeval2015}, 2016  \cite{semeval2016}) that laid influential foundations with widely-used product and service review ABSA benchmark datasets  \cite{RV21, RV19}. Nevertheless, the need for mining insights from opinions far exceeds this single domain. Many other areas, especially the public sector, also have an abundance of opinionated text data and can benefit from ABSA, such as helping policy-makers understand public attitudes and reactions towards events or changes  \cite{ref2}, improving healthcare services and treatments via patient experience and concerns in clinical visits, symptoms, drug efficacy and side-effects  \cite{239,ref105}, and guiding educators in meeting teacher and learner needs and improving their experience  \cite{OR15,8,55,72}. While the more general SA research has been applied to ``nearly every domain'' \cite[p.~1]{RV9}, this does not seem to be the case for ABSA. Chebolu et al. (2022) reviewed 62 public ABSA datasets released between 2004 and 2020 covering ``over 25 domains'' \cite[p.~1]{RV99}. However, 53 out of these 62 datasets were reviews of restaurants, hotels, and digital products; only five were not related to commercial products or services, and merely one was on the public sector domain (university reviews). 

The above-mentioned evidence raises questions: Will this dataset domain homogeneity be found with a larger sample of primary studies? Does this domain skewness reflect the concentration of ABSA research focus or merely the lack of dataset diversity? This motivated our \textbf{RQ1} (``\textit{To what extent is ABSA research and its dataset resources dominated by the commercial (especially the product and service review) domain? What proportion of ABSA research focuses on other domains and dataset resources?}'') Answers to these questions could inform and shape future ABSA research through individual research decisions and community resource collaboration. 

Second, there are many good survey papers on ABSA, most focused on introducing the methodological details of common ABSA subtasks and solutions (e.g.  \cite{RV1, RV10, RV21, RV22, RV24, RV12} or specific approaches such as DL methods for ABSA (e.g.  \cite{RV11, RV19, RV156, RV165, batch2_11, batch2_128, batch2_187}). We list these surveys in \textbf{Appendix \ref{A3_reviews}} as additional resources for the reader. Nonetheless, many of these reviews only explored each subtask and/or technique individually and often by iterating through reviewed studies, and few examined their combinations or changes over time and with quantitative evidence. For example, although the above-listed reviews \cite{RV11, RV19, RV156, RV165} reported the rise of DL approaches in ABSA similar to that of NLP as a whole, it is unclear whether ABSA research was also increasingly dominated by the attention mechanism from the Transformer architecture  \cite{AttentionPaper} and pre-trained large language models since 2018  \cite{OR20}, and if linguistic and traditional ML approaches were still active. In addition, most of these surveys used a smaller and selected sample that could not support conclusions on trends. As the field matures, we believe it is necessary and important to examine trends and matters outside the problem solution itself, so as to inform research decisions, identify issues, and call for necessary community awareness and actions. We thus proposed     \textbf{RQ2} (``\textit{What are the most common ABSA problem formulations via subtask combinations, and what proportion of ABSA studies only focus on a specific sub-task?}'') and \textbf{RQ3} (``\textit{What is the trend in the ABSA solution approaches over time? Are linguistic and traditional machine-learning approaches still in use?}'').

In order to identify patterns and trends for our RQs, a sufficiently sized representative sample and systematic approach are required. We chose to conduct an SLR, as this type of review aims to answer specific research questions from all available primary research evidence following well-defined review protocols  \cite{OR14}. Moreover, none of the existing SLRs on ABSA share the same focus and RQs as ours: Among the 192 survey/review papers obtained from four major digital database searches detailed in Section \ref{sec3_methods}, only eight were SLRs on ABSA, within which four focused on non-English language(s)  \cite{RV2,RV3,RV6,RV7}, two on specific domains (software development, social media)  \cite{RV5,RV8}, one on a single subtask  \cite{RV1}, and one mentioned ABSA subtasks as a side-note under the main topic of SA  \cite{RV4}. 

In summary, this review aims to address gaps in the ABSA literature. The high-level nature of our research questions is best answered through a large-scale SLR to provide solid evidence. The next section presents our SLR approach and sample.


\section{Methods}\label{sec3_methods}

Following the guidance of Kitchenham and Charters (2007)  \cite{OR14}, we conducted this SLR with pre-planned scope, criteria, and procedures highlighted below. The complete SLR methods and process are detailed in \textbf{Appendix \ref{appendix_B_method}}.

\subsection{Main Procedures}

For the main SLR sample, we sourced the primary studies in October 2022 from four major peer-reviewed digital databases:  ACM Digital Library, IEEE Xplore, Science Direct, and SpringerLink. First, we manually searched and extracted 4191 database  results without publication-year constraints. \textbf{Appendix \ref{sec_a1_database_search}} provides more details of the search strategies and results. Next, we applied the inclusion and exclusion criteria listed in \textbf{Table \ref{table2_criteria}} via automatic\footnote{Our PDF mining for automatic review screening code is available at \url{https://doi.org/10.5281/zenodo.12872948}} and manual screening steps and identified 519 in-scope peer-reviewed research publications for the review. The complete screening process, including that of the automatic screening, is described in \textbf{Appendix \ref{sec_a2_screening}}. We then manually reviewed the in-scope primary studies and recorded data following a planned scheme. Lastly, we checked, cleaned, and processed the extracted data and performed quantitative analysis against our RQs. 

\begin{table}[ht!]
\centering
  \caption{Inclusion and exclusion criteria used for this Systematic Literature Review (SLR)}
  \label{table2_criteria}

\begin{tabular}{p{0.3\linewidth}|p{0.6\linewidth}}
  
\toprule
    Inclusion Criteria & Exclusion Criteria\\ 
\midrule
    
\begin{enumerate} 
    \item	Published in the English language
    \item	Has both the sentiment analysis component and entity/aspect-level granularity
    \item	Focuses on text data
    \item	Is a primary study with quantitative elements in the ABSA approach
    \item	Contains original solutions or ideas for ABSA task(s)
    \item	Involves experiment and results on ABSA task(s)
\end{enumerate} 
&
\begin{enumerate}
    \item	The main text of the article is not in the English language
    \item	Missing either the sentiment analysis or entity/aspect-level granularity in the research focus
    \item	Only contains search keyword in the reference section
    \item	Only contains less than 5 ABSA-related search keywords outside the reference section 
    \item	Is not a primary study (e.g. review articles, meta-analysis)
    \item	Does not provide quantitative experiment results on ABSA task(s)
    \item	Has fewer than three pages
    \item	Contains multimodal (i.e. non-text) input data for ABSA task(s)
    \item	The research focus is not on ABSA task(s), even though ABSA models might be involved  (e.g. recommender system)
    \item	Focuses on transferring existing ABSA approaches between languages 
    \item	The ABSA tasks are integrated into a model built for other purposes, and there are no stand-alone ABSA method details and/or evaluation results
\end{enumerate} \\

\bottomrule
\end{tabular}
\end{table}

\subsection{Main SLR Sample Summary}

\begin{figure*}[ht!]
  \centering
  \includegraphics[width=\linewidth]{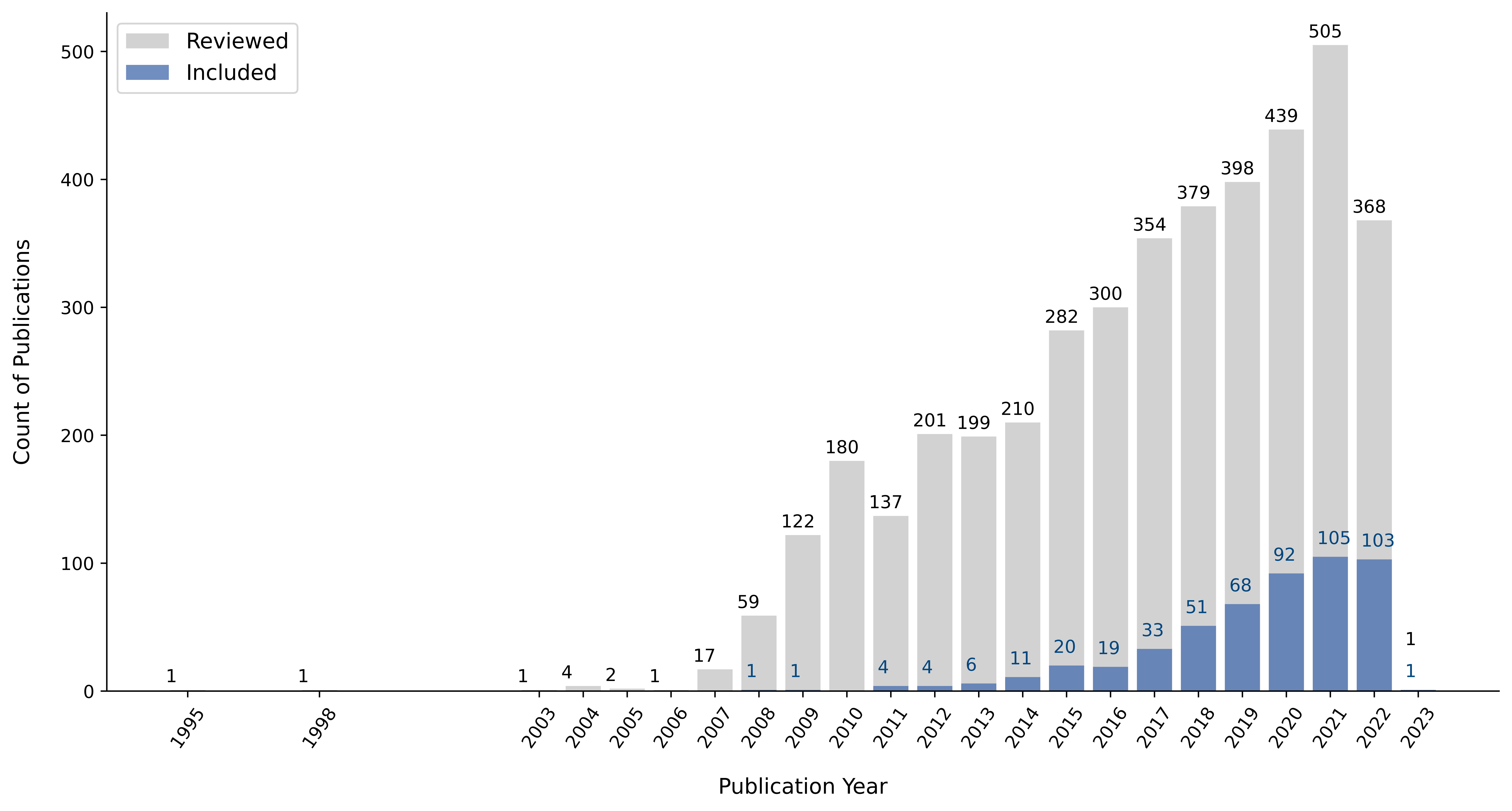}
  \caption{Number of Studies by Publication Year:  Total reviewed (N=4191) vs. Included (N=519)}
  \label{fig1}
\end{figure*}

\textbf{Figure~\ref{fig1}} shows the number of total reviewed vs. included studies across all publication years for the 4191 SLR search results. The search results include studies published between 1995 and 2023 (N=1), although all of the pre-2008 ones (2 from the 90s, 8 from 2003-2006, 17 from 2007) were not ABSA-focused and were excluded during automatic screening. The earliest in-scope ABSA study in the sample was published in 2008, followed by a very sparse period until 2013. The numbers of extracted and in-scope publications have both grown noticeably since 2014, a likely result of the emergence of deep learning approaches, especially sequence models such as RNNs  \cite{OR20,OR21}. We also present a breakdown of the included studies by publication year and type in \textbf{Figure~\ref{fig_c1}} in Appendix \ref{appendix_C_results}. 

\subsection{Note on ``domain'' mapping}

In order to answer RQ1, we made the distinction between ``research application domain'' (``research domain'' in short) and ``dataset domain'', and manually examined and classified each study and its datasets into domain categories. 

We considered each study's research domain to be ``non-specific'' unless the study mentioned a specific application domain or use case as its motivation. For the dataset domain, we examined each dataset used by our sample, standardised its name, and recorded the domain from which it was drawn/selected based on the description provided by the author or the dataset source webpage. Datasets without a specific domain (e.g. Twitter tweets crawled without a specific domain filter) were labelled as ``non-specific''.  

We then manually grouped the research and dataset domains into 19 common categories used for analysis. More details and examples on domain mapping are available in \textbf{Appendix \ref{sec_a3_domain_mapping}}.

\subsection{Phase-2 targeted review on in-context learning}

Additionally, generative ``foundation models'' \cite{foundationmodel}, defined as models with billions of parameters pre-trained on enormous general-purpose data and adaptable to diverse downstream NLP tasks, have become ubiquitous after our SLR data collection (e.g. ChatGPT \cite{refChatGPT}, released in November 2022). We use the term ``foundation models'' to distinguish them from the earlier pre-trained Large Language Models (LLMs) such as BERT \cite{refBERT}, BART \cite{BART}, and T5 \cite{T5}, which have relatively fewer parameters and typically require fine-tuning for task adaptation \cite{batch2_128}. These generative foundation models brought a new paradigm of ``In-context Learning'' (ICL) \cite[p.~4]{refGPT3}, where task adaptation can occur solely via conditioning the model on the text input instructions (``prompts'') with zero (``zero-shot ICL'') or few (``few-shot ICL'') examples and no model parameter changes \cite{refGPT3, incontextlearning}. To capture and analyse this new development while balancing feasibility and currency, we conducted a Phase-2 targeted review in July 2024. 

This Phase-2 targeted review focuses solely on the ICL implementations of pre-trained generative models for ABSA tasks, excluding those involving fine-tuning to draw a distinction from other non-ICL deep-learning approaches covered in the SLR. To extend the SLR sample beyond the original extraction time, we conducted a new database search\footnote{This new database search followed the same procedures and criteria as the SLR, except that we aborted the SpringerLink search due to persistent database interface search result navigation issues during our data collection period.} in July 2024 for studies published from 2022 onwards and removed the ones already included in the SLR sample. The new search results were screened using the SLR criteria described in \textbf{Table \ref{table2_criteria}} and then combined with the 519 SLR final samples. We then applied an additional filtering condition ``Gen-LLM'' to all the in-scope ABSA primary studies, which further selected publications with at least one occurrence of any of the following keywords outside the Reference section: ``generative'', ``in-context'', ``in context learning'', ``genai'', ``bart'', ``t5'', ``flan-t5'', ``gpt'', ``chatgpt'', ``llama'', and ``mistral''. With the help of our automatic screening pipeline detailed in \textbf{Appendix \ref{sec_a2_screening}}, we were able to efficiently auto-screen the new search results and re-screen the previous SLR sample for the "Gen-LLM" keywords in less than one hour.

In total, the new search yielded 271 additional in-scope ABSA primary studies from 4359 search results. After applying the ``Gen-LLM'' filtering condition to the combined 790 in-scope ABSA primary studies, we obtained 208 Phase-2 samples for manual review, which comprised 91 studies from the new search and 117 from the earlier SLR sample, ranging from 2008 to 2024. The Phase-2 targeted review results are presented in Section \ref{4.5_genLLM}. Unless specified otherwise, the results below only refer to those of the SLR.


\section{Results}\label{sec4_results}

This section presents the SLR results corresponding to each of the RQs:

\subsection{Results for RQ1}

\begin{quote} 
    RQ1. To what extent is ABSA research and its dataset resources dominated by the commercial (especially the product and service review) domain? What proportion of ABSA research focuses on other domains and dataset resources?
\end{quote}

To answer RQ1, we examined the distribution of reviewed studies by their research (application) domains, dataset domains, and the relationship between the two. From the 519 reviewed studies, we recorded 218 datasets, 19 domain categories (15 research domains and 17 dataset domains), and obtained 1179 distinct ``study-dataset'' pairs and 630 unique ``study \& dataset-domain'' combinations. The key results are summarised below and presented in \textbf{Table~\ref{table4_domains}} and \textbf{Figure~\ref{fig_b2}}. We also list the datasets used by more than one reviewed study in the Appendix \textbf{Table~\ref{tab:c5_dataset}}.

\begin{table}[ht!]
\centering
    \caption{Number of In-Scope Studies Per Each Research (Application) and Dataset Domain Category}
    \label{table4_domains}
\begin{tabular}%
{>{\raggedright\arraybackslash}p{0.28\linewidth}%
 >{\raggedleft\arraybackslash}p{0.12\linewidth}%
 >{\raggedleft\arraybackslash}p{0.12\linewidth}%
 >{\raggedleft\arraybackslash}p{0.12\linewidth}%
 >{\raggedleft\arraybackslash}p{0.12\linewidth}}
\toprule
\textbf{Domain} &
  \textbf{Count of Studies per Research Domain} &
  \textbf{\% of Studies per Research Domain} &
  \textbf{Count of Studies per Dataset Domain} &
  \textbf{\% of Studies per Dataset Domain} \\ 
\midrule
Non-specific                        & \textbf{339} & \textbf{65.32\%}  & 98            & 15.56\%           \\
Product/service review              & 126           & 24.28\%           & \textbf{447} & \textbf{70.95\%}  \\
Student feedback / education review & 12            & 2.31\%            & 19            & 3.02\%            \\
Politics/ policy-reaction           & 8             & 1.54\%            & 5             & 0.79\%            \\
Healthcare/ medicine                & 7             & 1.35\%            & 9             & 1.43\%            \\
Video/movie review                  & 6             & 1.16\%            & 19            & 3.02\%            \\
News                                & 5             & 0.96\%            & 8             & 1.27\%            \\
Finance                             & 5             & 0.96\%            & 4             & 0.63\%            \\
Research/academic reviews           & 3             & 0.58\%            & 3             & 0.48\%            \\
Disease                             & 3             & 0.58\%            & 3             & 0.48\%            \\
Employer review                     & 1             & 0.19\%            & 1             & 0.16\%            \\
Nuclear energy                      & 1             & 0.19\%            & 0             & 0.00\%            \\
Natural disaster                    & 1             & 0.19\%            & 1             & 0.16\%            \\
Music review                        & 1             & 0.19\%            & 1             & 0.16\%            \\
Multiple domain                     & 1             & 0.19\%            & 0             & 0.00\%            \\
Location review                     & 0             & 0.00\%            & 8             & 1.27\%            \\
Book review                         & 0             & 0.00\%            & 2             & 0.32\%            \\
Biology                             & 0             & 0.00\%            & 1             & 0.16\%            \\
Singer review                       & 0             & 0.00\%            & 1             & 0.16\%            \\
\midrule
\textbf{TOTAL}                      & \textbf{519}  & \textbf{100.00\%} & \textbf{630}  & \textbf{100.00\%} \\ 
\bottomrule
\end{tabular}
\end{table}

\begin{figure*}[!ht]
  \centering
  \includegraphics[width=\linewidth]{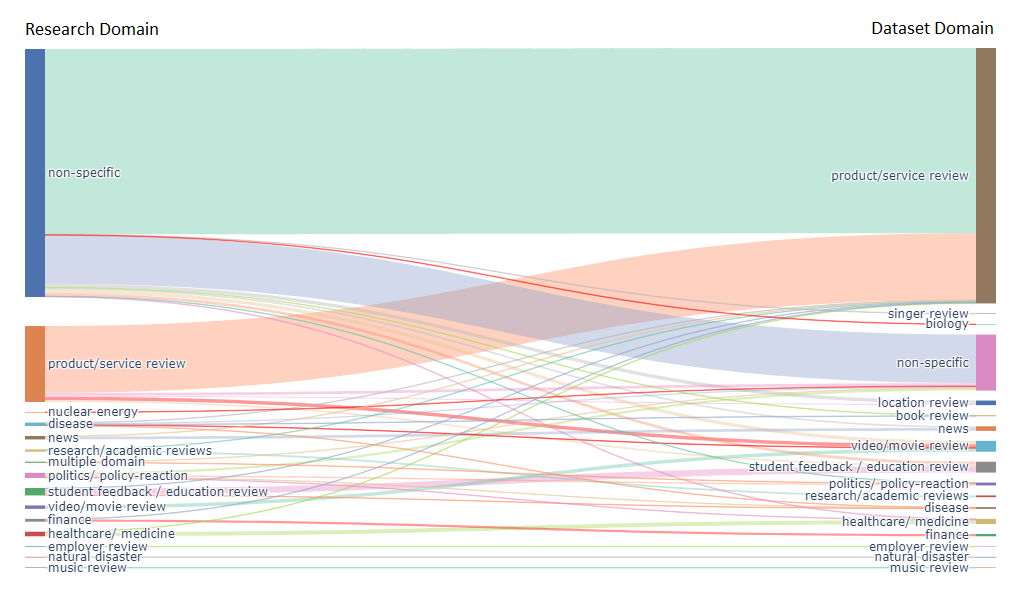}
  \caption{Distribution of Unique ``Study-Dataset'' Pairs (N=1179, with 519 studies and 218 datasets) by Research (Application) Domains (left) and Dataset Domains (right). Note: 1) The top flow visualises a mismatch between the two domains: the majority of studies without a specific research domain used datasets from the product/service review domain. 2) The disproportionately small number of samples in both domains that were neither ``non-specific'' nor ``product/service review''.}
  \label{fig_b2}
\end{figure*}

In summary, our results answer RQ1 by showing that: 1) The majority (65.32\%) of the reviewed studies were not for any specific application domain and only 24.28\% targeted ``product/service review''. 2) The dataset resources used in the sample were mostly domain-specific (84.44\%) and dominated by the ``product/service review'' datasets (70.95\%). 3) Both the research effort and dataset resources were scant in the non-commercial domains, especially the main public sector areas, with fewer than 13 studies across 14 years in each of the healthcare, policy, and education domains, where about half of the used datasets were created from scratch for the study. 

Beyond RQ1, 1) and 2) above also suggest a significant mismatch between the research and dataset domains as visualised in \textbf{Figure~\ref{fig_b2}}. Further, when filtering out datasets used by less than 10 studies, we discovered an alarming lack of dataset diversity as only 12 datasets remained, of which 10 were product/service reviews. When examining the three-way relationship among research domain, dataset domain, and dataset name, we further identified an over-representation (78.20\%) of the four SemEval restaurant and laptop review benchmark datasets. This is illustrated in \textbf{Figure~\ref{fig4}}.

\subsubsection{Detailed Results for RQ1}

\begin{figure*}[ht!]
  \centering
  \includegraphics[width=\linewidth]{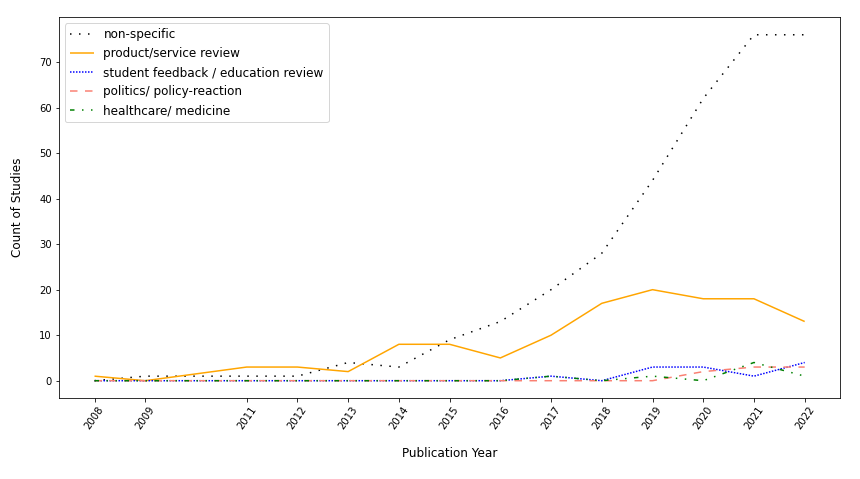}
  \caption{Number of In-Scope Studies by Research (Application) Domain and Publication Year (N=518). This graph excludes the one 2023 study (extracted in October 2022) to avoid trend confusion.}
  \label{fig3}
\end{figure*}

For research (application) domains indicated by the stated research use case or motivation, the majority (65.32\%, N=339) of the 519 reviewed studies have a ``non-specific'' research domain, followed by just a quarter (24.28\%, N=126) in the ``product/service review'' category. However, the number of studies in the rest of the research domains is magnitudes smaller in comparison, with only 12 studies (2.31\%) in the third largest category ``student feedback/education review'' since 2008, followed by 8 in Politics/policy-reaction (1.54\%), and only 7 in Healthcare/medicine (1.35\%). \textbf{Figure~\ref{fig3}} revealed further insights from the trend of research domain categories with five or more reviewed studies. Interestingly, ``product/service review'' has been a persistently major category over time, and has only been consistently taken over by ``non-specific'' since 2015. The sharp increase of domain-``non-specific'' studies since 2018 could be partly driven by the rise of pre-trained language models such as BERT and the greater sequence processing power from the Transformer architecture and the attention mechanism  \cite{OR20}, as more researchers explore the technicalities of ABSA solutions.

As to the dataset domains, \textbf{Table~\ref{table4_domains}} suggests that among the 630 unique ``study \& dataset-domain'' pairs, the majority (70.95\%, N=447) are in the ``product/service review'' category, followed by 15.56\% (N=98) in ``Non-specific''. The third place is shared by two magnitude-smaller categories: ``student feedback/ education review'' (3.02\%, N=19) and ``video/movie review'' (3.02\%, N=19). The numbers of studies with datasets from the Healthcare/medicine (1.43\%, N=9) and Politics/policy-reaction (0.79\%, N=5) domains were again single-digit. Moreover, nearly half of the unique datasets in the public domains were created by the authors for the first time: 5/9 in Healthcare/medicine, 2/4 in Politics/policy-reaction, and 8/12 in Student feedback/ Education review. 

\begin{figure*}[ht!]
  \centering
  \includegraphics[width=\linewidth]{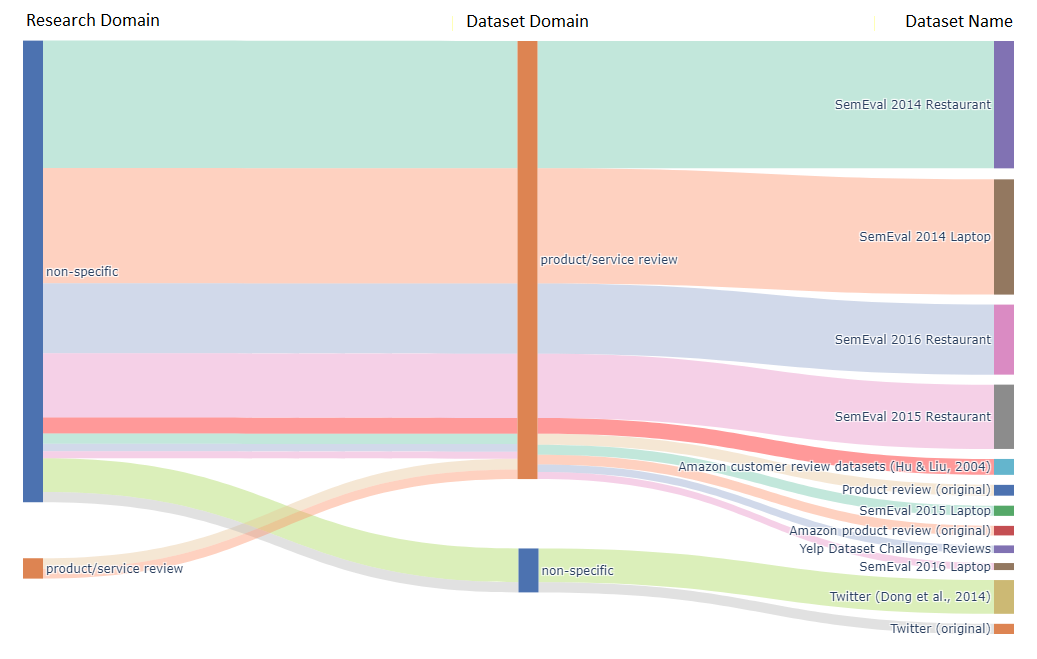}
  \caption{Number of Studies Per Each Research (Application) Domain (left), Dataset Domain (middle), and Dataset (right) Combination, filtered by Datasets Used by 10 or More In-Scope Studies (N=757). The three-way relationship highlights that not only did the majority of the sample studies with ``non-specific'' research domain use datasets from the `product/service review` domain, but their datasets were also dominated by only four SemEval datasets on two types of product and service reviews.}
  \label{fig4}
\end{figure*}

Furthermore, to understand the dataset diversity across samples and domains, we grouped the 1179 unique ``study-dataset'' pairs by ``research-domain, dataset-domain, dataset-name'' combinations and zoomed into the 757 entries with ten or more study counts each. As shown in \textbf{Table~\ref{table5_3way}} and illustrated in \textbf{Figure~\ref{fig4}}, among these 757 unique combinations, 95.77\% (N=725) are in the ``non-specific'' research domain, of which 90.48\% (N=656) used ``product/service review'' datasets. Most interestingly, these 757 entries only involve 12 distinct datasets of which 10 were product and service reviews, and 78.20\% (N=592) are taken up by the four SemEval datasets from the early pioneer work  \cite{semeval2014,semeval2015,semeval2016} mentioned in Subsection \ref{2.4_rationale}: SemEval 2014 Restaurant, SemEval 2014 Laptop (these two alone account for 50.33\% of all 757 entries), SemEval 2016 Restaurant, and SemEval 2015 Restaurant. This finding echos  \cite{OR17,RV99}: ``The SemEval challenge datasets … are the most extensively used corpora for aspect-based sentiment analysis''  \cite[~p.4]{RV99}. Meanwhile, the top dataset used under ``product/service review'' research and dataset domains is the original product review dataset created by the researchers.  \cite{RV99,ref14} provides a detailed introduction to the SemEval datasets.

\begin{table}[ht!]
\centering
    \caption{Number of Studies Per Each Research (Application) Domain, Dataset Domain, and Dataset Combination for All Datasets Used by Ten or More In-Scope Studies (N=757)}
    \label{table5_3way}
\begin{tabular}%
{>{\raggedright\arraybackslash}p{0.18\linewidth}%
 >{\raggedright\arraybackslash}p{0.18\linewidth}%
 >{\raggedright\arraybackslash}p{0.22\linewidth}%
 >{\raggedleft\arraybackslash}p{0.12\linewidth}%
 >{\raggedleft\arraybackslash}p{0.1\linewidth}}
\toprule
\textbf{Research Domain} & \textbf{Dataset Domain} & \textbf{Dataset}                                  & \textbf{Count of Studies} & \textbf{\% of Studies} \\ \midrule
non-specific   & product/service review & SemEval 2014 Restaurant        & 200          & 26.42\%           \\
non-specific   & product/service review & SemEval 2014 Laptop            & 181          & 23.91\%           \\
non-specific   & product/service review & SemEval 2016 Restaurant        & 110          & 14.53\%           \\
non-specific   & product/service review & SemEval 2015 Restaurant        & 101          & 13.34\%           \\
non-specific   & non-specific           & Twitter (Dong et al., 2014)    & 53           & 7.00\%            \\
non-specific             & product/service review  & Amazon customer review datasets (Hu \& Liu, 2004) & 25                        & 3.30\%                 \\
product/service review   & product/service review  & Product review (original)                         & 17                        & 2.25\%                 \\
non-specific   & non-specific           & Twitter (original)             & 16           & 2.11\%            \\
non-specific   & product/service review & SemEval 2015 Laptop            & 16           & 2.11\%            \\
product/service review   & product/service review  & Amazon product review (original)                  & 15                        & 1.98\%                 \\
non-specific   & product/service review & Yelp Dataset Challenge Reviews & 12           & 1.59\%            \\
non-specific   & product/service review & SemEval 2016 Laptop            & 11           & 1.45\%            \\
\midrule
\textbf{Total} & \textbf{}              & \textbf{}                      & \textbf{757} & \textbf{100.00\%} \\ 
\bottomrule
\end{tabular}
\end{table}

It is noteworthy that among the 519 reviewed studies, 20 focused on cross-domain or domain-agnostic ABSA, and 19 of them did not have a specific research application domain. However, while all 20 studies used multiple datasets, 17 solely involved the ``product/service review'' domain category by using reviews of restaurants and different products, and 14 used at least one SemEval dataset. The only three studies that went beyond the ``product/service review'' dataset domain added in movie reviews, singer reviews, and generic tweets.

\subsection{Results for RQ2}

\begin{quote} 
    RQ2. What are the most common ABSA problem formulations via subtask combinations, and what proportion of ABSA studies only focus on a specific sub-task?
\end{quote}

\begin{figure*}[ht!]
  \centering
  \begin{subfigure}[b]{0.45\textwidth}
        \raggedleft
        \includegraphics[height=0.43\textheight]{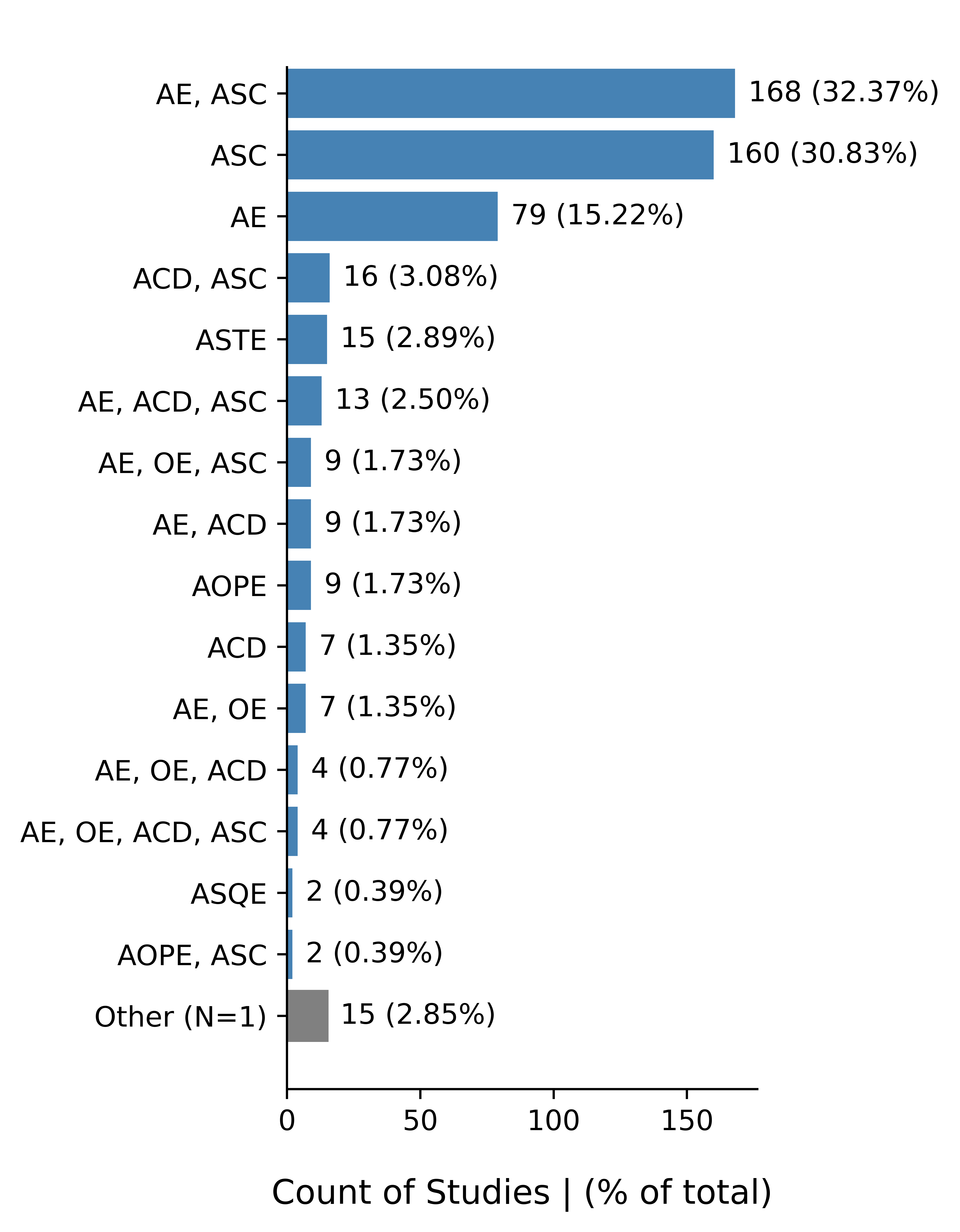}
        \caption{Number of Studies per ABSA Subtask Combinations (N=519)}
        \label{fig:fig7}
    \end{subfigure}
  \hfill
  \begin{subfigure}[b]{0.45\textwidth}
        \raggedright
        \includegraphics[height=0.43\textheight]{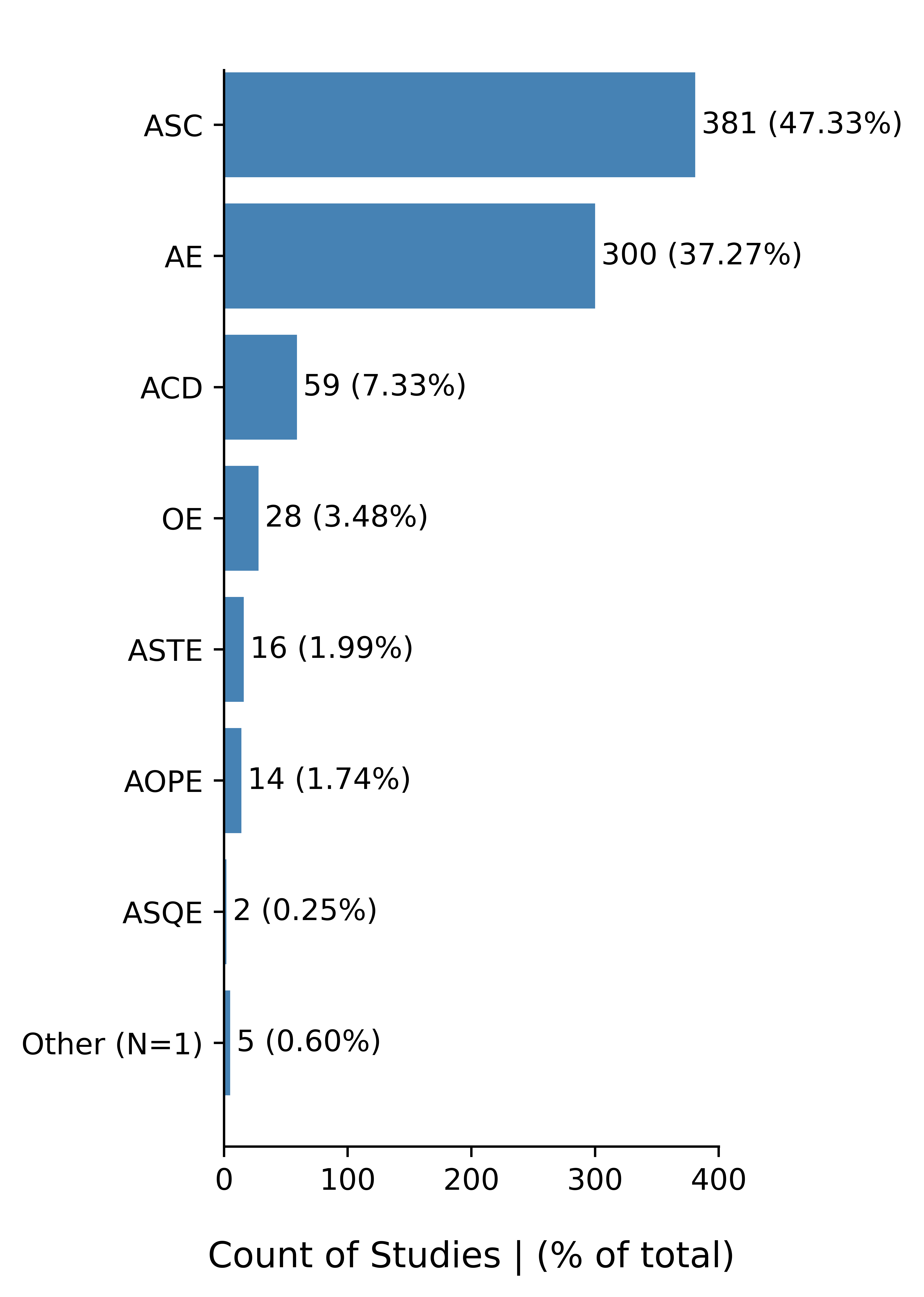}
        \caption{Number of Studies per Individual ABSA Subtask (N=805)}
        \label{fig:fig8}
    \end{subfigure}
    \caption{Number of Studies by ABSA Subtask}
    \label{fig:7-8}
\end{figure*}

For RQ2, we examined the 13 recorded subtasks and 805 unique ``study-subtask'' pairs to identify the most explored ABSA subtasks and subtask combinations across the 519 reviewed studies. As shown in \textbf{Figure~\ref{fig:fig7}}, 32.37\% (N=168) of the studies developed full-ABSA solutions through the combination of AE and ASC, and a similar proportion (30.83\%, N=160) focused on ASC alone, usually formulating the research problem as contextualised sentiment analysis with given aspects and the full input text. Only 15.22\% (N=79) of the studies solely explored the AE problem. This is consistent with the number of studies by individual subtasks shown in \textbf{Figure~\ref{fig:fig8}}, where ASC is the most explored subtask, followed by AE and ACD. 

\begin{figure*}[ht!]
  \centering
  \includegraphics[width=\linewidth]{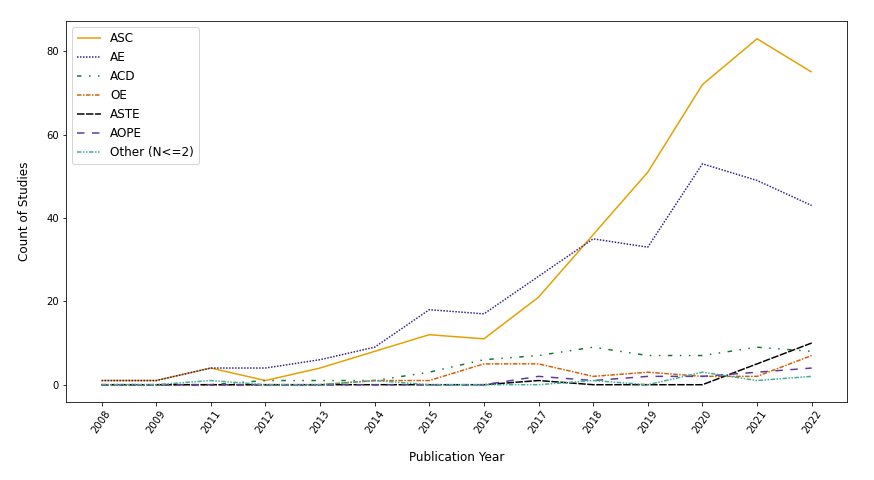}
  \caption{Distribution of Unique ``Study - ABSA Subtask'' Pairs by Publication Year (N=805).  
 This graph excludes the one 2023 study (extracted in October 2022) to avoid trend confusion.}
  \label{fig5}

\end{figure*}

Moreover, \textbf{Figure~\ref{fig5}} reveals a small but noticeable rise in composite subtask ASTE since 2020 (N=1, 5 and 10 in 2017, 2021, 2022) and a decline in ASC and AE around the same period. This could signify a problem formulation shift driven by deep-learning, especially multi-task learning methods for E2E ABSA. Our Phase-2 targeted review findings in \textbf{Subsection \ref{4.5_genLLM}} add more insights into this.

\subsection{Results for RQ3}

\begin{quote} 
    RQ3. What is the trend in the ABSA solution approaches over time? Are linguistic and traditional machine-learning approaches still in use?
\end{quote}

To answer RQ3, we examined the 519 in-scope studies along two dimensions, which we call ``paradigm'' and ``approach''. We use ``\textbf{paradigm}'' to indicate whether a study employed techniques along the supervised-unsupervised dimension and other types, such as reinforcement learning. We classify non-machine-learning approaches under the ``unsupervised'' paradigm, as our focus is on dataset and resource dependency. By ``\textbf{approach}'', we refer to the more specific type of techniques, such as deep learning (DL), traditional machine learning (traditional ML), linguistic rules (``rules'' for short), syntactic features and relations (``syntactics'' for short), lexicon lists or databases (``lexicon'' for short), and ontology or knowledge-driven approaches (``ontology'' for short).  

Overall, the results suggest that our samples are dominated by fully- (60.89\%) and partially-supervised (5.40\%) ML methods that are more reliant on annotated datasets and prone to their impact. As to ABSA solution approaches, the sample shows that DL methods have rapidly overtaken traditional ML methods since 2017, particularly with the prevalent RNN family (55.91\%) and its combination with the fast-surging attention mechanism (26.52\%). Meanwhile, traditional ML and linguistic approaches have remained a small but steady force even in the most recent years. Context engineering through introducing linguistic and knowledge features to DL and traditional ML approaches was very common. More detailed results and richer findings are presented below. 

\subsubsection{Paradigms}

\textbf{Table~\ref{table8_paradigm}} lists the number of studies per each of the main paradigms. Among the 519 reviewed studies, 66.28\% (N= 344) is taken up by those using somewhat- (i.e. fully-, semi- and weakly-) supervised paradigms that have varied levels of dependency on labelled datasets, where the fully-supervised ones alone account for 60.89\% (N=316). Only 19.65\% (N=102) of the studies do not require labelled data, which are mostly unsupervised (18.69\%, N=97). In addition, hybrid studies are the third largest group (14.07\%, N=73).

\begin{table}[ht!] 
\centering
    \caption{Number of Studies by Paradigm (N=519). \\
    Note: The unsupervised category includes non-ML approaches}
    \label{table8_paradigm}
\begin{tabular}%
{>{\raggedright\arraybackslash}p{0.3\linewidth}%
 >{\raggedleft\arraybackslash}p{0.2\linewidth}%
 >{\raggedleft\arraybackslash}p{0.15\linewidth}}
\toprule

\textbf{Paradigm}      & \textbf{Count of Studies} & \textbf{\% of Studies} \\ \midrule
Fully-supervised       & 316                       & 60.89\%                \\
Unsupervised           & 97                        & 18.69\%                \\
Hybrid                 & 73                        & 14.07\%                \\
Semi-supervised        & 22                        & 4.24\%                 \\
Weakly-supervised      & 6                         & 1.16\%                 \\
Reinforcement learning & 3                         & 0.58\%                 \\
Self-supervised        & 2                         & 0.39\%                 \\
\midrule
\textbf{TOTAL}         & \textbf{519}              & \textbf{100.00\%}      \\

\bottomrule
\end{tabular}
\end{table}

\begin{figure*}[ht!]
  \centering
  \begin{subfigure}[b]{0.45\textwidth}
        \raggedleft
        \includegraphics[height=0.43\textheight]{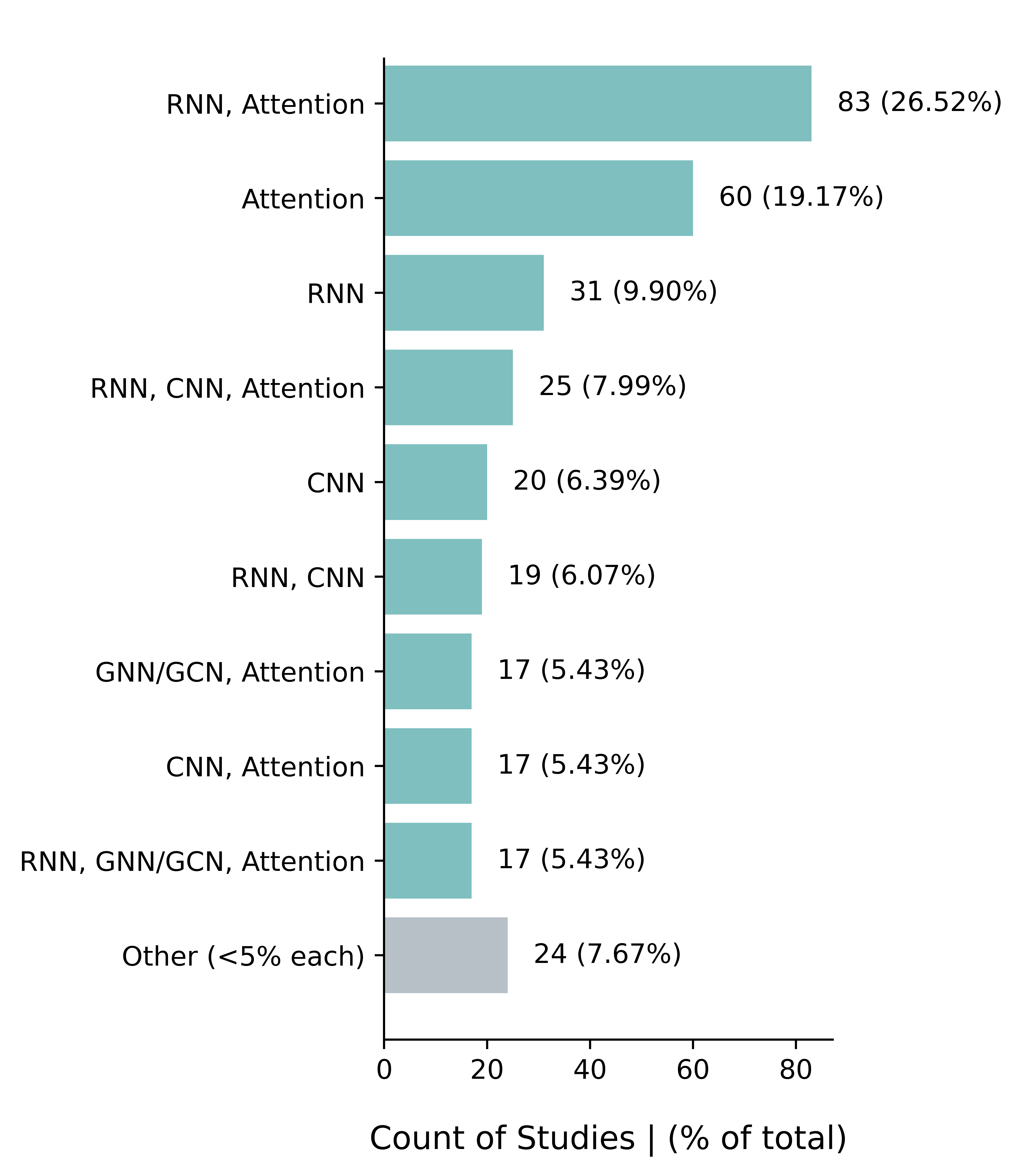}
        \caption{Number of Studies by DL Approaches
(N=313)}
        \label{fig:fig10}
    \end{subfigure}
  \hfill
  \begin{subfigure}[b]{0.45\textwidth}
        \raggedright
        \includegraphics[height=0.43\textheight]{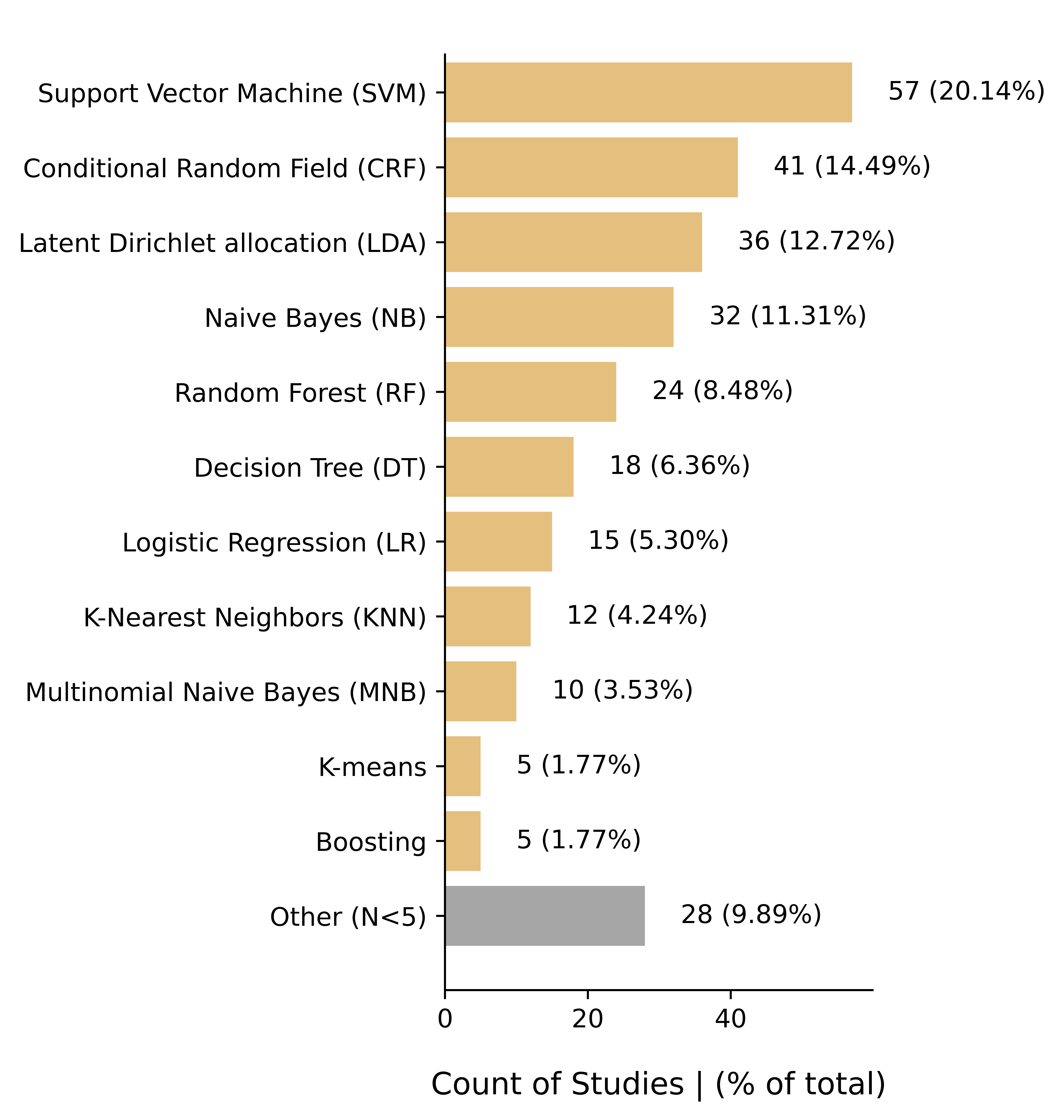}
        \caption{Number of Studies by Traditional ML Approaches (N=283)}
        \label{fig:fig11}
    \end{subfigure}
    \caption{Number of Studies using DL and Traditional ML approaches}
    \label{fig:9-10}
\end{figure*}

We further analysed the approaches under each paradigm and focused on three for more details: deep learning (DL), traditional machine learning (ML), and Linguistic and Statistical Approaches. The results are detailed below and presented in \textbf{Figure~\ref{fig:9-10}} and \textbf{Tables~\ref{table10_paradigm_approach}, ~\ref{table12_lingstat}}.

\begin{table}[ht!] 
\centering
    \caption{Number of Studies by Paradigm and Approaches (N=519)}
    \label{table10_paradigm_approach}
\begin{tabular}%
{>{\raggedright\arraybackslash}p{0.22\linewidth}%
>{\raggedright\arraybackslash}p{0.39\linewidth}%
 >{\raggedleft\arraybackslash}p{0.12\linewidth}%
 >{\raggedleft\arraybackslash}p{0.1\linewidth}}
 
\toprule
\textbf{Paradigm}        & \textbf{Approaches}                              & \textbf{Count of Studies} & \textbf{\% of Studies} \\ \midrule
supervised             & DL                                                   & 149                & 47.15\%                \\
                       & Syntactics, DL                                       & 53                 & 16.77\%                \\
                       & DL, traditional ML                                   & 23                 & 7.28\%                 \\
                       & Other (\textless{}5\% each)                          & 91                 & 28.80\%                \\
Total                  &                                                      & \textbf{316}       & \textbf{100.00\%}      \\
\midrule
semi-supervised        & DL                                                   & 7                  & 31.82\%                \\
                       & Syntactics, Lexicon, traditional ML                  & 3                  & 13.64\%                \\
                       & traditional ML                                       & 2                  & 9.09\%                 \\
                       & Rules, Syntactics, Lexicon                           & 2                  & 9.09\%                 \\
                       & Other (\textless{}5\% each)                          & 8                  & 36.36\%                \\
Total                  &                                                      & \textbf{22}        & \textbf{100.00\%}               \\
\midrule
weakly-supervised      & DL, traditional ML                                   & 2                  & 33.33\%                \\
                       & traditional ML                                       & 1                  & 16.67\%                \\
                       & DL                                                   & 1                  & 16.67\%                \\
                       & Syntactics, DL                                       & 1                  & 16.67\%                \\
                       & Syntactics, Ontology, DL                             & 1                  & 16.67\%                \\
                       & Other (\textless{}5\% each)                          & 0                  & 0.00\%                 \\
Total                  &                                                      & \textbf{6}         & \textbf{100.00\%}               \\
\midrule
self-supervised        & DL                                                   & 2                  & 100.00\%               \\
                       & Other (\textless{}5\% each)                          & 0                  & 0.00\%                 \\
Total                  &                                                      & \textbf{2}         & \textbf{100.00\%}               \\
\midrule
unsupervised           & Rules, Syntactics, Lexicon                           & 24                 & 24.74\%                \\
                       & Rules, Syntactics, Lexicon, Ontology                 & 15                 & 15.46\%                \\
                       & Rules, Syntactics                                    & 8                  & 8.25\%                 \\
                       & traditional ML                                       & 7                  & 7.22\%                 \\
                       & Other (\textless{}5\% each)                          & 43                 & 44.33\%                \\
Total                  &                                                      & \textbf{97}                 & \textbf{100.00\%}               \\
\midrule
hybrid                 & Rules, Syntactics, traditional ML                    & 6                  & 8.22\%                 \\
                       & Rules, Syntactics, Lexicon, traditional ML           & 6                  & 8.22\%                 \\
                       & traditional ML                                       & 5                  & 6.85\%                 \\
                       & Syntactics, traditional ML                           & 4                  & 5.48\%                 \\
                       & Rules, Syntactics, Lexicon                           & 4                  & 5.48\%                 \\
                       & Rules, Syntactics, Lexicon, Ontology, traditional ML & 4                  & 5.48\%                 \\
                       & Other (\textless{}5\% each)                          & 44                 & 60.27\%                \\
Total                  &                                                      & \textbf{73}                 & \textbf{100.00\%}               \\
\midrule
reinforcement learning & DL                                                   & 1                  & 33.33\%                \\
                       & Syntactics, DL                                       & 1                  & 33.33\%                \\
                       & Other (\textless{}5\% each)                          & 1                  & 33.33\%                \\
Total         & \textbf{}                                            & \textbf{3}         & \textbf{100.00\%}      \\ 
\midrule
\textbf{TOTAL}         & \textbf{}                                    & \textbf{519}         & \textbf{100.00\%}  \\    
\bottomrule
\end{tabular}
\end{table}

\subsection{Approaches} \label{4.4_approaches}

As shown in \textbf{Figure~\ref{fig:fig10}} and \textbf{Table~\ref{table10_paradigm_approach}}, among the 519 reviewed studies, 60.31\% (N=313) employed DL approaches, and 30.83\% (N=160) are DL-only. The DL-only approach is particularly prominent among fully-supervised (47.15\%, N=149) and semi-supervised (31.82\%, N=7) studies. Supplementing DL with syntactical features is also the second most popular approach in fully-supervised studies (16.77\%, N=53). 

\begin{itemize}[leftmargin=*]
    \item[] \textbf{1) DL Approaches} 
\end{itemize}

\textbf{Figure~\ref{fig:fig10}} suggests that the 313 studies involving DL approaches are dominated by Recurrent Neural Network (RNN)-based solutions (55.91\%, N=175), of which nearly half used a combination of RNN and the attention mechanism (26.52\%, N=83),  followed by attention-only (19.17\%, N=60) and RNN-only (9.90\%, N=31) models. The RNN family mainly consists of Long Short-Term Memory (LSTM), Bidirectional LSTM (BiLSTM), and Gated Recurrent Unit (GRU). These neural-networks are featured by sequential processing that captures temporal dependencies of text tokens, and can thus incorporate surrounding text as context for prediction \cite{RV11, batch2_187}. On the other hand, the sequential nature poses challenges with parallelisation and the exploding and vanishing gradient problems associated with long sequences \cite{AttentionPaper, RV11}. Although LSTM and GRU can mitigate these issues somewhat through cell state and memory controls, efficiency and long-dependency challenges still hinder their performance \cite{AttentionPaper, RV11, batch2_187}. The attention mechanism complements RNNs by dynamically updating weights across the input sequence based on each element's relevance to the current task, and thus guides the model to focus on the most relevant elements \cite{AttentionPaper}. 

In addition, convolutional and graph-neural approaches (e.g. Convolutional Neural Networks (CNN), Graph Neural Networks (GNN), Graph Convolutional Networks (GCN)) also play smaller but noticeable roles in DL-based ABSA studies. While CNN was commonly used as an alternative to the sequence models such as RNNs \cite{RV11,259,400}, the graph-based networks (GNN, GCN) were mainly used to model the non-linear relationships such as external conceptual knowledge (e.g. \cite{150}) and syntactic dependency structures (e.g. \cite{6,87,253}) that are not well captured by the sequential networks like RNNs and the flat structure of the attention modules. As a result, they inject richer context into the overall learning process \cite{152,238,399}.

\begin{figure*}
  \centering
  \includegraphics[width=\linewidth]{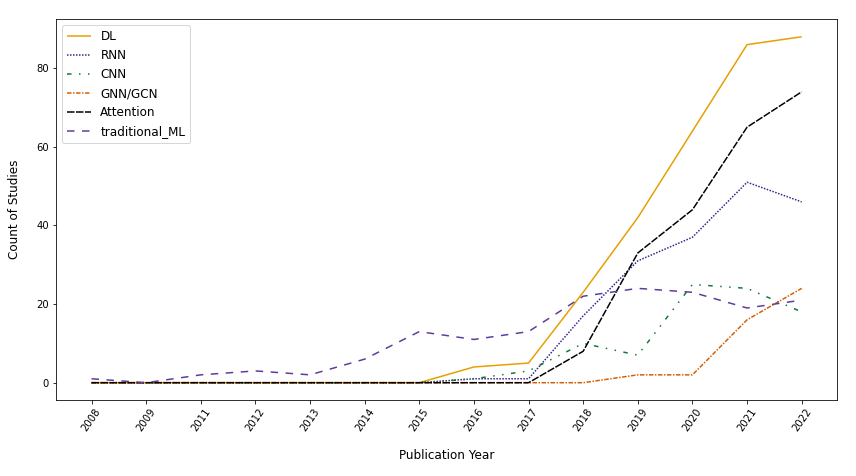}
  \caption{Distribution of Studies Per Method by Publication Year (N=1017 with 519 unique studies).  
 This graph excludes the one 2023 study (extracted in October 2022) to avoid trend confusion.}
  \label{fig6}
\end{figure*}

\textbf{Figure~\ref{fig6}} depicts the trend of the main approaches across the publication years. We excluded the one study pre-published for 2023 to avoid confusing trends. It is clear that DL approaches have risen sharply and taken dominance since 2017, mainly driven by the rapid growth in RNN- and attention-based studies. This coincides with the appearance of the Transformer architecture in 2017 \cite{AttentionPaper} and the resulting pre-trained models such as BERT \cite{refBERT} that were a popular embedding choice to be used alongside RNNs in DL and hybrid approaches (e.g., \cite{27,74}). GNN/GCN-based approaches remain small in number but have noticeable growth since 2020 (N=2, 2, 16, 24 in each of 2019-2022, respectively), suggesting an increased effort to dynamically integrate relational context into the learning process within the DL framework. 

\begin{itemize}[leftmargin=*]
    \item[] \textbf{2) Traditional ML Approaches} 
\end{itemize}

Interestingly, as shown in \textbf{Figure~\ref{fig6}} traditional ML approaches remain a steady force over the decades despite the rapid rise of DL methods. \textbf{Table~\ref{table10_paradigm_approach}} and \textbf{Figure~\ref{fig:fig11}} provide some insight into this: Among the 519 reviewed studies, while 60.31\% employed DL approaches as mentioned in the previous sub-section, over half (54.53\%, N=283) also included traditional ML approaches, with the top 3 being Support Vector Machine (SVM; 20.14\%, N=57), Conditional Random Field (CRF; 14.49\%, N=41), and Latent Dirichlet allocation (LDA; 12.72\%, N=36). \textbf{Table~\ref{table10_paradigm_approach}} suggests that among the major paradigms, traditional ML were often used in combination with DL approaches for fully-supervised studies (7.28\%, N=23), and along with linguistic rules, syntactic features, and/or lexicons and ontology in hybrid studies (27.40\%, N=20). Across all paradigms, traditional ML-only approaches are relatively rare (max N=7).

\begin{itemize}[leftmargin=*]
    \item[] \textbf{3) Linguistic and Statistical Approaches} 
\end{itemize}

    While \textbf{Table~\ref{table10_paradigm_approach}} illustrates the prevalence of fusing ML approaches with linguistic and statistical features or modules, there were 67 studies (12.91\% out of the total 519) on pure linguistic or statistical approaches. As shown in \textbf{Tables~\ref{table12_lingstat}}, although small in number, these non-ML approaches have persisted over time. The most popular combination (34.33\%, N=23) was rules built on syntactic features (e.g. POS tags and dependency parse trees) and used along lexicon resources (e.g. domain-specific aspect lists, SentiWordNet\footnote{\url{https://github.com/aesuli/SentiWordNet}}, MPQA\footnote{\url{https://mpqa.cs.pitt.edu/}}). A typical example is using POS tags and/or lexicon resources to narrow the scope of aspect or opinion term candidates, applying further rules based on POS tags or dependency relations for AE or OE, and/or using lexicon resources for candidate pruning, categorisation, or sentiment labelling (e.g. \cite{56,252,452}). The second top combination (23.88\%, N=16) is the above-mentioned one plus ontology (e.g. domain-specific ontology, ConceptNet\footnote{\url{https://conceptnet.io/}}, WordNet\footnote{\url{https://wordnet.princeton.edu/}}) to bring in external knowledge of concepts and relations (e.g. \cite{330,352}). Pure statistical methods were relatively rare (5.97\%, N=4), and mainly included frequency-based methods such as N-gram and TF-IDF, and other statistical modelling methods that were not commonly seen in the ML field.


\begin{table}[ht!]
\centering
    \caption{Number of Studies by Pure Linguistic or Statistical Approaches and Publication Year (N=67)}
    \label{table12_lingstat}
    
\begin{tabular}%
{>{\raggedright\arraybackslash}p{0.4\linewidth}%
>{\raggedleft\arraybackslash}p{0.05\linewidth}%
>{\raggedleft\arraybackslash}p{0.05\linewidth}%
>{\raggedleft\arraybackslash}p{0.05\linewidth}%
>{\raggedleft\arraybackslash}p{0.05\linewidth}|%
>{\raggedleft\arraybackslash}p{0.05\linewidth}%
>{\raggedleft\arraybackslash}p{0.08\linewidth}}

\toprule
 
\textbf{Approach}                       & \textbf{2011-2013} & \textbf{2014-2016} & \textbf{2017-2019} & \textbf{2020-2022} & \textbf{Total} & \textbf{Total\%}  \\ \midrule
Syntactics, Rules, Lexicon              & 1                  & 8                  & 8                  & 6                  & 23             & 34.33\%           \\
Syntactics, Rules, Lexicon, Ontology    & 1                  & 3                  & 9                  & 3                  & 16             & 23.88\%           \\
Syntactics, Rules                       &                    & 2                  & 4                  & 5                  & 11             & 16.42\%           \\
Statistical                             & 2                  & 1                  &                    & 1                  & 4              & 5.97\%            \\
Syntactics, Rules, Lexicon, Statistical & 1                  & 1                  & 1                  &                    & 3              & 4.48\%            \\
Syntactics, Rules, Statistical          &                    &                    & 2                  & 1                  & 3              & 4.48\%            \\
Lexicon                                 &                    &                    &                    & 1                  & 1              & 1.49\%            \\
Rules                                   &                    & 1                  &                    &                    & 1              & 1.49\%            \\
Rules, Lexicon                          &                    &                    & 1                  &                    & 1              & 1.49\%            \\
Rules, Lexicon, Statistical             &                    &                    &                    & 1                  & 1              & 1.49\%            \\
Syntactics, Lexicon                     &                    &                    &                    & 1                  & 1              & 1.49\%            \\
Syntactics, Lexicon, Statistical        &                    &                    & 1                  &                    & 1              & 1.49\%            \\
Syntactics, Statistical                 & 1                  &                    &                    &                    & 1              & 1.49\%            \\
\midrule
\textbf{TOTAL}                          & \textbf{6}         & \textbf{16}        & \textbf{26}        & \textbf{19}        & \textbf{67}    & \textbf{100.00\%} \\ \bottomrule
\end{tabular}
\end{table}

\subsection{ICL and generative approach in ABSA - Phase-2 targeted review results} \label{4.5_genLLM}

ICL is a subgenre of the DL approach. However, we discuss the relevant results in this separate subsection due to the Phase-2 review’s more focused sample and finer granularity. Despite the trending popularity of the ICL approach in NLP research and applications since 2022 \cite{incontextlearning}, our results suggest that the ABSA research community is just beginning to explore it with caution. Among the 208 ABSA studies from 2008 to 2024 containing at least one occurrence of the “Gen-LLM” keywords, only five (all published in 2024) applied ICL to both composite and traditional ABSA tasks. All of these studies were exploring the performance of foundation models via ICL against other approaches, rather than focusing on an ICL ABSA solution.  \textbf{Table \ref{table13_ICL}} summarises the models, ABSA tasks, and key findings from these studies. Overall, four of the five studies found that zero-shot and even 5-shot ICL on foundation models (mainly GPTs) could not reach the performance of fine-tuned or fully trained DL models, especially those leveraging pre-trained LLMs to fine-tune a contextual-embedding.

\begin{table}[ht!]
\centering
  \caption{Studies with In-context Learning (ICL) Approach on ABSA Tasks (N=5 out of 208 samples from 2008-2024) \\ 
  Note: The Non-ICL Models were trained or fine-tuned unless specified otherwise}
  \label{table13_ICL}

\begin{tabular}%
{>{\raggedright\arraybackslash}p{0.1\linewidth}%
>{\raggedright\arraybackslash}p{0.08\linewidth}%
>{\raggedright\arraybackslash}p{0.12\linewidth}%
>{\raggedright\arraybackslash}p{0.15\linewidth}%
>{\raggedright\arraybackslash}p{0.12\linewidth}%
>{\raggedright\arraybackslash}p{0.3\linewidth}
}

\toprule
\textbf{Paper} &
  \textbf{Task} &
  \textbf{Non-ICL Models} &
  \textbf{ICL Models} &
  \textbf{ICL approach} &
  \textbf{Result} \\ \midrule
 &
   &
   &
   &
   &
   \\
Zhou et al. (2024) \cite{batch2_04} &
  ASQE &
  RoBERTa &
  ChatGPT &
  5-shot ICL &
  ChatGPT performed worse than almost all non-ICL methods with about 20-30\% lower micro-F1 \\ \midrule
 &
   &
   &
   &
   &
   \\
Liu et al. (2024) \cite{batch2_01} &
  ASTE &
  BERT &
  ChatGPT &
  Zero-shot, 5-shot ICL &
  ChatGPT 5-shot ICL performed better than 0-shot ICL, but was still around 20 \% lower in F1 score than the main method \\ \midrule
 &
   &
   &
   &
   &
   \\
Su et al. (2024) \cite{batch2_09} &
  ASC &
  T5 &
  Llama2-7b-chat, Llama2-13b-chat, ChatGPT-3.5 &
  N.A. &
  ChatGPT 3.5 performed slightly better than the Llama-2 models across datasets, but was still up to about 20\% lower than the main model in accuracy and F1 \\ \midrule
 &
   &
   &
   &
   &
   \\
Amin et al. (2024) \cite{batch2_07} &
  AE, ASC, OE &
  LSTM, RoBERTa (not fine-tuned) &
  GPT 3.5-Turbo, GPT-4 &
  Zero-shot &
  On AE and ASC tasks, RoBERTa performed significantly better than the two GPT models with up to about 20\% higher accuracy.    GPT 3.5-Turbo performed better than GPT-4 on AE and ASC tasks, and outperformed all the other models with the OE task with up to 15\% higher accuracy \\ \midrule
 &
   &
   &
   &
   &
   \\
Mughal et al. (2024) \cite{batch2_11} &
  ASC, ACSA &
  LSTM, Flan-T5, DeBERTa, PaLM &
  GPT 3.5-Turbo, PaLM-bison &
  Zero-shot &
  PaLM showed the best overall performance on ASC and ASCA tasks in terms of accuracy and F1, closely followed by fine-tuned DeBERTa. However, PaLM had the lowest accuracy and F1 scores with the multi-aspect multi-sentiment (MAMS) dataset, with 21-48\% lower accuracy and F1 than fine-tuned DeBERTa \\ 
  
  \bottomrule

\end{tabular}
\end{table}

In addition, we identified an emerging trend by examining the Phase-2 review non-ICL samples: Those employing fine-tuned generative LLMs mostly formulated the ABSA tasks as Sequence-to-Sequence (Seq2Seq) text generation problems, with a particular focus on composite tasks such as ASTE and ASQE. As shown in  \textbf{Table \ref{table14_GenLLM}}, within the 208 samples, a total of 18 studies (all from the new search) published in 2022-2024 applied pre-trained generative LLMs with fine-tuning. The majority of these studies used models based on T5 (N=9) and BART (N=5) with the full Transformer \cite{AttentionPaper} encoder-decoder architecture, followed by encoder-only (N=3, BERT and RoBERTa \cite{ROBERTA}) and decoder-only (N=1, GPT-2 \cite{refGPT2}) models. All but two of these 18 studies were on composite ABSA tasks, mainly ASTE and ASQE. Moreover, two studies (\cite{batch2_16,batch2_22}) also leveraged the generative capability of these LLMs to augment training data to enrich the fine-tuned embedding. 

Compared with this Seq2Seq generation approach, the common applications of pre-trained LLMs in earlier studies from the main SLR sample often formulate the ABSA task as a classification problem \cite{batch2_128}. These studies mostly use encoder-only LLMs for their pre-trained representations to fine-tune a contextual embedding \cite{batch2_128}, which is then connected to other context-injection or relationship-learning modules and a classifier output layer. For instance, Zhang et al. (2022) \cite{74} employed pre-trained BERT with BiLSTM, a Feed-forward Neural Network (FFNN), and CRF. Li et al. (2021) \cite{27} used pre-trained BERT as an encoder and a decoder featuring a GRU. In contrast, the Seq2Seq generative approach can be illustrated by the signature ``Generative Aspect-based Sentiment analysis (GAS)'' proposed by Zhang et al. (2021) \cite{GAS}, which leveraged the LLM's pre-trained and fine-tuned encoder module for context-aware embedding and used the fine-tuned decoder module to generate text representations of the label sets (e.g., triplets) or as annotations next to the original input text \cite{GAS,batch2_128}.

\begin{table}[ht!]
\centering
\begin{threeparttable}[b]
  \caption{Studies with Fine-tuned Generative Large Language Models (LLMs) on ABSA Tasks (N=18 out of 208 samples from 2008-2024)}
  \label{table14_GenLLM}

  \renewcommand{\arraystretch}{1.5} 

\begin{tabular}%
{>{\raggedright\arraybackslash}p{0.28\linewidth}%
>{\raggedright\arraybackslash}p{0.3\linewidth}%
>{\raggedright\arraybackslash}p{0.28\linewidth}
}

\toprule
\textbf{Paper}                                              & \textbf{Task}   & \textbf{GenAI Model}                    \\ \midrule
Hoang et al. (2022) \cite{batch2_44}      & ASQE            & BART                                    \\
Kang et al. (2022) \cite{batch2_49}       & OE              & BART                                    \\
Gong \& Li (2022) \cite{batch2_87}        & ASTE            & BERT                                    \\
William \& Khodra (2022) \cite{batch2_17} & ASTE            & T5                                      \\
Li et al. (2023) \cite{batch2_25}         & ASQE            & BART (with BERT for upstream embedding) \\
Li et al. (2023) \cite{batch2_27}         & ASQE            & BART                                    \\
Suchrady \& Purwarianti (2023) \cite{batch2_03} &
  ASTE, AOPE, AE, OE, UABSA \tnote{1} &
  mT5 \tnote{2} \\
Yu et al. (2023) \cite{batch2_20} &
  AESC \tnote{3}, AOPE, ASTE &
  BART + GAT (Graph Attention Networks) \\
Yan et al. (2023) \cite{batch2_02}        & ASC,ASTE        & T5                                      \\
Yu et al. (2023) \cite{batch2_16} &
  AE, OE, ASQE &
  T5 (data generator + self-training) + BERT pair-classifier as descriminator \\
Lee \& Kim (2023) \cite{batch2_19}        & ASTE, ASQE, ASC & T5                                      \\
Liu et al. (2024) \cite{batch2_01}        & ASTE            & BERT + GPN (Global Pointer Network)     \\
Lee et al. (2024) \cite{batch2_14}        & AE, ASC         & GPT-2 fine-tuned by LoRA                \\
Dang et al. (2024) \cite{batch2_12} &
  AE, OE, ASC, ACD, AOPE, ACSA, ASTE, ACSD, ASQE &
  mT5, ViT5 \tnote{2} \\
Su et al. (2024) \cite{batch2_09}         & ASC             & T5                                      \\
Zhou et al. (2024) \cite{batch2_04}       & ASQE            & RoBERTa                                 \\
Wang et al. (2024) \cite{batch2_15} &
  AE, OE, ACD, ASC, ACSA, AOPE, ASTE, ASQE, ASPE \tnote{4}, CSPE \tnote{5} &
  T5 \\
Zhang et al. (2024) \cite{batch2_22}      & ASQE            & T5                                      \\

  \bottomrule

\end{tabular}

\begin{tablenotes}
\item[1] UABSA: Unified Aspect-based Sentiment Analysis. 
\item[2] mT5, ViT5: multi-lingual variants of T5.
\item[3] AESC: Aspect Term Extraction and Sentiment Classification.
\item[4] ASPE: Aspect Sentiment Pair Extraction.
\item[5] CSPE: Category Sentiment Pair Extraction. 

\end{tablenotes}
\end{threeparttable}
\end{table}


\section{Discussion}\label{sec5_discussion}

This review was motivated by the literature gap in capturing trends in ABSA research to answer higher-level questions beyond technical details, and the concern that the domain-dependent nature could predispose ABSA research to systemic hindrance from a combination of resource-reliant approaches and skewed resource domain distribution. By systematically reviewing the two waves of 727 in-scope primary studies published between 2008 and 2024, our quantitative analysis results identified trends in ABSA solution approaches, confirmed the above-mentioned concern, and provided detailed insights into the relevant issues. In this section, we examine the primary findings, share ideas for future research, and reflect on the limitations of this review.

\subsection{Significant Findings and Trends}

\subsubsection{The Out-of-sync Research and Dataset Domains}

Under RQ1, we examined the distributions of and relationships between our sample’s research (application) domains and dataset domains. The results showed strong skewness in both types of domains and a significant mismatch between them: While the majority (65.32\%, N=339) of the 519 studies did not aim for a specific research domain, a greater proportion (70.95\%, N=447) used datasets from the ``product/service review'' domain. A closer inspection of the link between the two domains revealed a clear mismatch: Among the 757 unique ``research-domain, dataset-domain, dataset-name'' combinations with ten or more studies:  90.48\% (N=656) of the studies in the ``non-specific'' research domain (95.77\%, N=725) used datasets from the ``product/service review'' domain. This suggests that the lack of non-commercial-domain datasets could have forced generic technical studies to use benchmark datasets from a single popular domain. Given ABSA problem's domain-dependent nature, this could have indirectly hindered the solution development and evaluation across domains. 

The results also showed that the other important and prevalent ABSA application domains such as education, medicine/healthcare, and public policy, were clearly under-researched and under-resourced. Among the reviewed samples from these three public-sector domains, about half of their datasets were created for the studies by their authors, indicating a lack of public dataset resources, hence the cost and challenge of developing ABSA research in these areas. As a likely consequence, even the most researched domain among these three had only 12 studies (2.31\% out of 519) since 2008. The dataset resource scarcity in these public sector domains deserves more research community attention and support, especially given these domains' overall low research resources vs. the high cost and domain knowledge required for quality data annotation. In particular, for domains such as ``Student feedback/education review” that often face strict data privacy and consent restrictions, it is crucial that the ABSA research community focus on creating ethical and open-access datasets to leverage community resources. 

\subsubsection{The Dominance and Limitations of the SemEval Datasets}

The results under RQ1 also revealed further issues with dataset diversity, even within the dominant ``product/service review'' domain. Out of the 757 unique ``research-domain, dataset-domain, dataset-name'' combinations with ten or more studies, 78.20\% (N=592) are taken up by the four popular SemEval datasets: The SemEval 2014 Restaurant and Laptop datasets alone account for 50.33\% of all 757 entries, and the other two (SemEval 2015 and 2016 Restaurant).

The level of dominance of the SemEval datasets is alerting, not only because of their narrow domain range, but also for the inheritance and impact of the SemEval datasets’ limitations. Several studies (e.g.  \cite{RV99,OR17,OR18,6}) suggest that these datasets fail to capture sufficient complexity and granularity of the real-world ABSA scenarios, as they primarily only include single-aspect or multi-aspect-but-same-polarity sentences, and thus mainly reflect sentence-level ABSA tasks and ignored subtasks such as multi-aspect multi-sentiment ABSA. The experiment results from  \cite{OR17,OR18,6} consistently showed that all 35 ABSA models (including those that were state-of-the-art at the time) (9 in  \cite{OR17}, 16 in  \cite{OR18}, 10 in  \cite{6}) that were trained and performed well on the SemEval 2014 ABSA datasets showed various extents of performance drop (by up to 69.73\% in  \cite{OR17}) when tested on same-source datasets created for more complex ABSA subtasks and robustness challenges. Given that the SemEval datasets are heavily used as both training data and ``benchmark'' to measure ABSA solution performance, their limitations and prevalence are likely to form a self-reinforcing loop that confines ABSA research. To break free from this dataset-performance self-reinforcing loop, it is critical that the ABSA research community be aware of this issue, and develop and adopt datasets and practices that are robustness-oriented, such as the automatic data-generation framework and the resulting Aspect Robustness Test Set (ARTS) developed by  \cite{OR17} for probing model robustness in distinguishing target and non-target aspects, and the Multi-Aspect Multi-Sentiment (MAMS) dataset created by  \cite{OR18} to reflect more realistic challenges and complexities in aspect-term sentiment analysis (ATSA) and aspect-category sentiment analysis (ACSA) tasks.

\subsubsection{The Reliance on Labelled-Datasets}

The domain and dataset issues discussed above would not be as problematic if most ABSA studies employed methods that are dataset-agnostic. However, our results under RQ3 show the opposite. Only 19.65\% (N=102, with 97 being unsupervised) of the 519 reviewed studies do not require labelled data, whereas 66.28\% (N= 344) are somewhat-supervised, and fully-supervised studies alone account for 60.89\% (N=316). 

As demonstrated in Section \ref{2.3_domainDependency}, the domain can directly affect whether a chunk of text is considered an aspect or the relevant sentiment term, and plays a crucial role in contextual inferences such as implicit aspect extraction and multi-aspect multi-sentiment pairing. The domain knowledge reflected via ABSA labelled datasets can further shape the linguistic rules, lexicons, and knowledge graphs for non-machine-learning approaches; and define the underpinning feature space, representations, and acquired relationships and inferences for trained machine-learning models. When applying a solution built on datasets from a domain that is very remote from or much narrower than the intended application domain,  it is predictable that the solution performance would be capped at subpar and even fail at more context-heavy tasks  \cite{ref3,127,150,111,17,OR1,RV9}. Thus, domain transfer is crucially necessary for balancing the uneven ABSA research and resource distributions across domains. However, our finding that 17 out of the 20 reviewed cross-domain or domain-agnostic ABSA studies solely used datasets from the ``product/service review'' domain raised questions about these approaches' generalisability and robustness in other domains, as well as whether such dataset choices became another reinforcement of concentrating research effort and benchmarks within this one dominant domain.

The rapid rise of deep learning (DL) in ABSA research could further add to the challenge of overcoming the negative impact of this domain mismatch and dataset limitations via the non-linear multi-layer dissemination of bias in the representation and learned relations, thus making problem-tracking and solution-targeting difficult. In reality, of the 519 reviewed studies, 60.31\% (N=313) employed DL approaches, and nearly half (47.15\%, N=149) of the fully-supervised studies and 30.83\% (N=160) of all reviewed studies were DL-only. 

Moreover, RNN-based solutions dominate the DL approaches (55.91\%, N=175), mainly with the RNN-attention combination (26.52\%, N=83) and RNN-only (9.90\%, N=31) models. RNN and its variants such as LSTM, BiLSTM, and GRU are known for their limitations in capturing long-distance relations due to their sequential nature and the subsequent memory constraints \cite{AttentionPaper, RV11}. Although the addition of the attention mechanism enhances the model's focus on more important features such as aspect terms \cite{AttentionPaper, RV11}, traditional attention weights calculation struggles with multi-word aspects or multi-aspect sentences \cite{RV11, Fan2018MultigrainedAN}. In addition, whilst 16.77\% (N=53) of the fully-supervised studies introduced syntactical features to their DL solutions, additional features also increased the input size. According to  \cite{OR9}, sequential models, even the state-of-the-art LLMs, showed impaired performance as the input grew longer and could not always benefit from additional features.

\subsubsection{The Potential of Generative LLMs and Foundation Models}

Lastly, the Phase-2 targeted review highlights the ABSA community's caution toward the direct adoption of generative foundation models, with only five out of 208 recent studies testing the ICL approach and most yielding subpar results compared to other methods. However, most of these studies only tested zero-shot instructions with simple model settings. It is worth further exploring the potential of foundational models and ICL in ABSA by focusing more on instruction and example engineering, model parameter optimisation, and task re-formulation \cite{incontextlearning}. 

On the other hand, the fine-tuning of smaller generative LLMs has seen increasing adoption through the ``ABSA as Seq2Seq text generation'' approach, demonstrating promising task performance. Although this generative approach can incorporate data augmentation and self-training to reduce reliance on labelled datasets, the cost of fine-tuning, the need for labelled base data, and the domain-transfer problem remain significant challenges \cite{batch2_128}. In this context, the task adaptability and multi-domain pre-trained knowledge of foundation models could provide potential solutions.

As Zhang et al. (2022) \cite{batch2_128} noted, progress in applying pre-trained LLMs and foundation models to ABSA could be impeded by dataset resources constraints. To match the parameter size of these models, more diverse, complex, and larger datasets are required for effective fine-tuning or comprehensive testing. In low-resource domains where dataset resources are already limited, this requirement could further complicate the adoption of these technologies \cite{batch2_187}.

\subsection{Ideas for Future Research}

Overall, by adopting a ``systematic perspective, i.e., model, data, and training''  \cite[~p.28]{6} combined with a quantitative approach, we identified high-level trends unveiling the development and direction of ABSA research, and found clear evidence of large-scale issues that affect the majority of the existing ABSA research. The skewed domain distributions of resources and benchmarks could also restrict the choice of new studies. On the other hand, this evidence also highlights areas that need more attention and exploration, including: ABSA solutions and resource development for the less-studied domains (e.g. education and public health), low-resource and/or data-agnostic ABSA, domain adaptation, alternative training schemes such as adversarial (e.g. \cite{6, 210}) and reinforcement learning (e.g. \cite{446, 404}), and more effective feature and knowledge injection. Future research could contribute to addressing these issues by focusing on ethically producing and sharing more diverse and challenging datasets in minority domains such as education and public health, improving data synthesis and augmentation techniques, exploring methods that are less data-dependent and resource-intensive, and leveraging the rapid advancements in pre-trained LLMs and foundation models.

In addition, our results also revealed emerging trends and new ideas. The relatively recent growth of end-to-end models and composite ABSA subtasks provide opportunities for further exploration and evaluation. The fact that hybrid approaches with non-machine-learning techniques and non-textual features remain steady forces in the field after nearly three decades suggests valuable characteristics that are worth re-examining under the light of new paradigms and techniques. Moreover, the small number of Phase-2 samples using ICL and fine-tuning generative LLM approaches may suggest that we have only captured early adopters. More thorough exploration of these approaches and continued tracking of their development alongside other methods are necessary to understand how the ABSA community can leverage the resources and capabilities embedded within LLMs and foundation models.

Lastly, it is crucial that the community invest in solution robustness, especially for machine-learning approaches  \cite{OR17,OR18,6}. This could mean critical examination of the choice of evaluation metrics, tasks, and benchmarks, and being conscious of their limitations vs. the real-world challenges. The ``State-Of-The-Art'' (SOTA) performance based on certain benchmark datasets should never become the motivation and holy grail of research, especially in fields like ABSA where the real use cases are often complex and even SOTA models do not generalise far beyond the training datasets. More attention and effort should be paid to analysing the limitations and mistakes of ABSA solutions, and drawing from the ideas of other disciplines and areas to fill the gaps.  

\subsection{Limitations}

We acknowledge the following limitations of this review: First, our sample scope is by no means exhaustive, as it only includes primary studies from four peer-reviewed digital databases and only those published in the English language. Although this can be representative of a core proportion of ABSA research, it does not generalise beyond this without assumptions. The ``peer-reviewed'' criteria also meant that we overlooked preprint servers such as arXiv.org that more closely track the latest development of ML and NLP research. Second, no search string is perfect. Our database search syntax and auto-screening keywords represent our best effort in capturing ABSA primary studies, but may have missed some relevant ones, especially with the artificial ``total pages $<$ 3'' and ``total keyword (except SA, OM) outside Reference $<$ 5'' exclusion criteria. Moreover, our search completeness might have been affected by the performance of the database search engines. This is evidenced by the significant number of extracted search results that were entirely irrelevant to the search keywords, as well as our abandonment of the 2024 SpringerLink search due to interface issues. Enhancements in digital database search capabilities could significantly improve the effectiveness and reliability of future literature review studies, particularly SLRs. Third, we may have missed datasets, paradigms, and approaches that are not clearly described in the primary studies, and our categorisation of them is also subject to the limitations of our knowledge and decisions. Future review studies could consider a more innovative approach to enhance analytical precision and efficiency, such as applying ABSA and text summarisation alongside the screening and reviewing process. Fourth, we did not compare solution performance across studies due to the review focus, sample size, and the variability in experimental settings across studies. Evaluating the effectiveness of comparable methods and the suitability of evaluation metrics would enhance our findings and offer more valuable insights.  


\section{Conclusion}\label{sec6_conclusion}

ABSA research is riding the wave of the explosion of online digital opinionated text data and the rapid development of NLP resources and ideas. However, its context- and domain-dependent nature and the complexity and inter-relations among its subtasks pose challenges to improving ABSA solutions and applying them to a wider range of domains. In this review, we systematically examined existing ABSA literature in terms of their research application domain, dataset domain, and research methodologies. The results suggest a number of potential systemic issues in the ABSA research literature, including the predominance of the ``product/service review'' dataset domain among the majority of studies that did not have a specific research application domain, coupled with the prevalence of dataset-reliant methods such as supervised machine learning. We discussed the implication of these issues to ABSA research and applications, as well as their implicit effect in shaping the future of this research field through the mutual reinforcement between resources and methodologies. We suggested areas that need future research attention and proposed ideas for exploration.

\newpage









\textbf{Authors' contributions} Y. C. H designed, conducted, and wrote this review. P. D., J. W., and K. T. guided the design of the review methodology and review protocol, and reviewed and provided feedback on the manuscript.

\section*{Declarations}\label{sec_declaration}


\textbf{Competing Interests}
The authors have no competing interests as defined by Springer, or other interests that might be perceived to influence the results and/or discussion reported in this paper.

\clearpage
\newpage

\onecolumn
\appendix

\setcounter{figure}{0}
\setcounter{table}{0}

\renewcommand\thefigure{\thesection.\arabic{figure}}
\renewcommand\thetable{\thesection.\arabic{table}}
\renewcommand\theHtable{\thesection.\thetable}
\renewcommand{\thefigure}{\thesection.\arabic{figure}} 
\renewcommand{\theHfigure}{\thesection.\arabic{figure}} 

\centering \section*{APPENDICES}


\section{Aspect-Based Sentiment Analysis (ABSA)} \label{appendix_A_absa}

\subsection{\textbf{Definition and examples}} \label{A1_absa}

\textbf{Aspect-Based Sentiment Analysis (ABSA) }is a sub-domain of fine-grained SA  \cite{1}. ABSA focuses on identifying the sentiments towards specific entities or their attributes/ features called \textbf{aspects}  \cite{1,2}. An aspect can be explicitly expressed in the text (\textbf{explicit aspect}) or absent from the text but implied from the context (\textbf{implicit aspects})  \cite{RV1,374}. Moreover, the aspect-level sentiment could differ across aspects and be different from the overall sentiment of the sentence or the document (e.g.  \cite{2,5,13}). Some studies further distinguish aspect into \textbf{aspect term} and \textbf{aspect category}, with the former referring to the aspect expression in the input text (e.g. ``pizza''), and the latter a latent construct that is usually a high-level category across aspect terms (e.g. ``food'') that are either identified or given  \cite{174,195}. 

The following examples illustrate the ABSA terminologies:

\begin{quote}
    \textbf{Example 1} (from a restaurant review\footnote{\url{https://alt.qcri.org/semeval2014/task4/}}): ``\textbf{The restaurant was expensive, but the menu was great.}'' 
    
    This sentence has one explicit aspect ``menu'' (sentiment term: ``great'', sentiment polarity: positive),  one implicit aspect ``price'' (sentiment term: ``expensive'', sentiment polarity: negative). Depending on the target/given categories, the aspects can be further classified into categories, such as ``menu'' into ``general'' and ``price'' into ``price''.\\

    \textbf{Example 2} (from a laptop review\footnote{\url{https://alt.qcri.org/semeval2015/task12/}}): ``\textbf{It is extremely portable and easily connects to WIFI at the library and elsewhere.}'' 

    This sentence has two implicit aspects: ``portability'' (sentiment term: ``portable'', sentiment polarity: positive), ``connectivity'' (sentiment term: ``easily'', sentiment polarity: positive). The aspects can be further classified into categories, such as both under ``laptop'' (as opposed to ``software'' or ``support'').\\

    \textbf{Example 3} (text from a course review): ``\textbf{It was too difficult and had an insane amount of work, I wouldn’t recommend it to new students even though the tutorial and the lecturer were really helpful}.'' 

    The two explicit aspects in Example 3 are ``tutorial'' and ``lecturer'' (sentiment terms: ``helpful'', polarities: positive). The implicit aspects are ``content'' (sentiment term: ``too difficult'', sentiment polarity: negative), ``workload'' (sentiment term: ``insane amount'', sentiment polarity: negative), and ``course'' (sentiment term: ``would not recommend'', sentiment polarity: negative).   An illustration of aspect categories would be assigning the aspect ``lecturer'' to the more general category ``staff'' and ``tutorial'' to the category ``course component''.
\end{quote}

As demonstrated above, the fine granularity makes ABSA more targetable and informative than document- or sentence-level SA. Thus, ABSA can precede downstream applications such as attribute weighting in overall review ratings (e.g.  \cite{57}), aspect-based opinion summarisation (e.g.  \cite{205,232,473}, and automated personalised recommendation systems (e.g.  \cite{274,452}).

Compared with document- or sentence-level SA, while being the most detailed and informative, ABSA is also the most complex and challenging  \cite{19}. The most noticeable challenges include the number of ABSA subtasks, their interrelations and context dependencies, and the generalisability of solutions across topic domains.

\subsection{\textbf{ABSA Subtasks}} \label{A2_subtasks}

A full ABSA solution has more subtasks than coarser-grained SA. The most fundamental ones  \cite{13,19,70,253} include:

\begin{quote} 
\textbf{Aspect (term) Extraction/Identification (AE)}, which has a slight variation in meaning depending on the overall ABSA approach. Some authors (e.g.  \cite{14,171,84}) consider AE as identifying the attribute or entity that is the target of an opinion expressed in the text and sometimes call it ``opinion target extraction''  \cite{198}. In these cases, opinion terms were often identified in order to find their target aspect terms. Others (e.g.  \cite{2,20,70,159,237}) define AE as identifying the key or all attributes of entities mentioned in the text. \textbf{Implicit-Aspect Extraction (IAE)} is often mentioned as a task by itself due to its technical challenge.  

\textbf{Opinion (term) Extraction/Identification (OE)}, which relates to identifying the ``opinion terms'' or the sentiment expression of a specific entity/aspect (e.g.  \cite{13,132,203,253}). In Example 1 above, an OE task would extract the sentiment terms ``great'' (associated with the aspect term ``menu'') and ``expensive'' (associated with the implicit aspect ``price'').

\textbf{Aspect-Sentiment Classification (ASC)}, which refers to obtaining the sentiment polarity category (e.g. negative, neutral, positive, conflict) or sentiment score (e.g. 1 to 5 or -1 to 1 along the scale from negative to positive) associated with a given aspect or aspect category (e.g.  \cite{2,504,516}). This is often done via evaluating the associated opinion term(s), and sentiment lexicon resources such as the SentiWordNet  \cite{OR22} and SenticNet  \cite{OR23} can be used to assign polarity scores  \cite{504}. Sentiment scores can be further aggregated across opinion terms for the same aspect, or across aspect terms to generate higher-level ratings, such as aspect-category ratings within or across documents  \cite{504,516}. 

As an extension of AE, some studies also involve \textbf{Aspect-Category Detection (ACD)} and \textbf{Aspect Category Sentiment Analysis (ACSA)} when the focus of sentiment analysis is on (often pre-defined) latent topics or concepts and requires classifying aspect terms into categories  \cite{207}.  
\end{quote}

Traditional full ABSA solutions often perform the subtasks in a pipeline manner  \cite{21,71} using one or more of the linguistic (e.g. lexicons, syntactic rules,  dependency relations), statistical (e.g. n-gram, Hidden Markov Model (HMM)), and machine-learning approaches  \cite{RV1,RV5,330}. For instance, for AE and OE, some studies used linguistic rules and sentiment lexicons to first identify opinion terms and then the associated aspect terms of each opinion term, or vice versa (e.g.  \cite{127,239}), and then moved on to ASC or ACD using a supervised model or unsupervised clustering and/or ontology  \cite{452,504}. Hybrid approaches are common given the task combinations in a pipeline.

With the rise of multi-task learning and deep learning  \cite{421}, an increasing number of studies explore ABSA under an \textbf{End-to-end (E2E)} framework that performs multiple fundamental ABSA subtasks in one model to better capture the inter-task relations \cite{batch2_01}, and some combine them into a single composite task  \cite{19,21,OR1}. These composite tasks are most commonly formulated as a sequence- or span-based tagging problem  \cite{19,21,71}. The most common composite tasks are:  \textbf{Aspect-Opinion Pair Extraction (AOPE)}, which directly outputs \{aspect, opinion\} pairs from text input  \cite{71,87,139} such as ``\textlangle menu, great\textrangle'' from Example 1;  \textbf{Aspect-Polarity Co-Extraction (APCE)}  \cite{19,OR6}, which outputs \{aspect, sentiment polarity\} pairs such as ``\textlangle menu, positive\textrangle'';  \textbf{Aspect-Sentiment Triplet Extraction (ASTE)}   \cite{19,21,152,253}, which outputs \{aspect, opinion, sentiment category\} triplets, such as ``\textlangle menu, great, positive\textrangle''; and \textbf{Aspect-Sentiment Quadruplet Extraction/Prediction (ASQE/ASQP)} \cite{74,372,OR25,ref-PARAPHRASE} that outputs \{aspect, opinion, aspect category, sentiment category\} quadruplets, such as ``\textlangle menu, great, general, positive\textrangle''. 

\subsection{\textbf{Other ABSA Reviews}} \label{A3_reviews}

As this review focuses on trends instead of detailed solutions and methodologies, we refer interested readers to existing review papers that provide comprehensive and in-depth summaries of common ABSA subtask solutions and approaches, for example:

\begin{itemize}

    \item \textbf{Explicit and implicit AE}: \quad Rana and Cheah (2016) \cite{RV21}, Ganganwar and Rajalakshmi (2019) \cite{RV24}, Soni and Rambola (2022) \cite{RV22}, Maitama et al. (2020) \cite{RV1}

    \item \textbf{Deep learning (DL) methods for ABSA}: \quad Do et al. (2019) \cite{RV19}, Liu et al. (2020) \cite{RV11}, Wang et al. (2021) \cite{RV156}, Chen and Fnu (2022) \cite{RV165}, Zhang et al. (2022) \cite{batch2_128},  Mughal et al. (2024) \cite{batch2_11}.  
    
    Specifically: 
   
    \begin{itemize} 

        \item[*] \textbf{DL methods for ASC}: \quad Zhou et al. (2019) \cite{RV12}, Satyarthi and Sharma (2023) \cite{batch2_187} 
        
        \item[*] \textbf{E2E ABSA, composite tasks, and pre-trained Large Language Models (LLMs) in ABSA}: \quad Zhang et al. (2022) \cite{batch2_128} provided a comprehensive review and shared extensive reading lists and dataset resource links via \url{https://github.com/IsakZhang/ABSA-Survey}. Mughal et al. (2024) \cite{batch2_11} introduced common benchmark datasets, including more challenging ones for composite ABSA tasks. They also reviewed and tested the ABSA task performance of representative RNN-based models and pre-trained LLMs. 
    
    \end{itemize}
    
    \item \textbf{Multimodal ABSA}: \quad Zhao et al. (2024) \cite{batch2_145}

\end{itemize}

\newpage
\section{Full SLR Methodology}\label{appendix_B_method} 

This section provides a complete, detailed description of the SLR methodology and procedures.

\subsection{Research identification} \label{sec_a1_database_search}

To obtain the files for review, we conducted database searches between 24-25 October 2022, when we manually queried and exported a total of 4191 research papers’ PDF and BibTeX (or the equivalent) files via the web interfaces of four databases. \textbf{Table~\ref{table_a1_database}} details the search string, search criteria, and the PDF files exported from each database. 

\begin{table}[ht!]
\centering
\small
  \caption{Digital databases and search details used for this Systematic Literature Review (SLR)}
  \label{table_a1_database}
  
\begin{tabular}{p{0.1\linewidth}|p{0.22\linewidth}|p{0.23\linewidth}|p{0.06\linewidth}|p{0.14\linewidth}}

\toprule
  \textbf{Database} &
  \textbf{Search String} &
  \textbf{Search Criteria} &
  \textbf{PDFs exported} &
  \textbf{Type of exported publications} 
\\
\midrule
ACM Digital Library 
&
((``aspect based'' OR ``aspect-based'') AND (sentiment OR extraction OR extract OR mining)) OR ``opinion mining'' 
&
\begin{itemize}[leftmargin=*]
    \item No year filter 
    \item Search scope = Full article
    \item Content Type = Research Article
    \item Media Format = PDF
    \item Publications = Journals OR Proceedings OR Newsletters
\end{itemize} 
& 
1514 
&
\begin{itemize}[leftmargin=*] 
    \item 201 articles
    \item 1283 conference papers
    \item 30 newsletters
\end{itemize} 
\\ \midrule
IEEE Xplore 
&
((``aspect based'' OR ``aspect-based'') AND (sentiment OR extraction OR extract OR mining)) OR ``opinion mining''
&
\begin{itemize}[leftmargin=*] 
    \item Year filter = 2004-2022  (pilot search suggested that 1995-2003 results were irrelevant)
    \item Search scope = Full article
    \item Publications = Not ``Books''
\end{itemize} 
& 
1639
&
\begin{itemize}[leftmargin=*] 
    \item 165 articles
    \item 1445 conference papers
    \item 29 magazine pieces
\end{itemize} 
\\ \midrule
Science Direct 
&
(``aspect based'' OR ``aspect-based'') AND (sentiment OR extraction OR extract OR mining)) OR ``opinion mining''
&
\begin{itemize}[leftmargin=*] 
    \item No year filter
    \item Search scope = Title, abstract or author-specified keywords
    \item Publications = Research Articles only
\end{itemize} 
& 
497
&
\begin{itemize}[leftmargin=*] 
    \item 497 articles
\end{itemize} 
\\ \midrule
SpringerLink
&
(``aspect based'' OR ``aspect-based'') AND (sentiment OR extraction OR extract OR mining)) OR ``opinion mining''
&
\begin{itemize}[leftmargin=*]
    \item No year filter
    \item English results only
    \item Publications = Article, Conference Paper
\end{itemize} 
& 
541
&
\begin{itemize}[leftmargin=*]  
    \item 218 articles
    \item 323 conference papers 
\end{itemize} \\
\bottomrule
\end{tabular}
\end{table}

Given the limited search parameters allowed in these digital databases, we adopted a ``search broad and filter later'' strategy. These database search strings were selected based on pilot trials to capture the ABSA topic name, the relatively prevalent yet unique ABSA subtask term (``extraction''), and the interchangeable use between ABSA and opinion mining; while avoiding generating false positives from the highly active, broader field of SA. The ``filter later'' step was carried out during the ``selection of primary studies'' stage introduced in the next section, which aimed at excluding cases where the keywords are only mentioned in the reference list or sparsely mentioned as a side context, and opinion mining studies that were at document or sentence levels. 

\subsection{Selection of primary studies} \label{sec_a2_screening}

After obtaining the 4191 initial search results, we conducted a pilot manual file examination of 100 files to refine the pre-defined inclusion and exclusion criteria. We found that some search results only contained the search keywords in the reference list or Appendix, which was also reported in  \cite{OR9}. In addition, there are a number of papers that only mentioned ABSA-specific keywords in their literature review or introduction sections, and the studies themselves were on coarser-grained sentiment analysis or opinion mining. Lastly, there were instances of very short research reports that provided insufficient details of the primary studies. Informed by these observations, we refined our inclusion and exclusion criteria to those in \textbf{Table~\ref{table2_criteria}} in Section \ref{sec3_methods}. Note that we did not include popularity criteria such as citation numbers so we can better identify novel practices and avoid mainstream method over-dominance introduced by the citation chain  \cite{OR13}.

To implement the inclusion and exclusion criteria, we first applied PDF mining to automatically exclude files that meet the exclusion criteria, and then refined the selection with manual screening under the exclusion and inclusion criteria. Both of these processes are detailed below. Our PDF mining for automatic review screening code is also available at \url{https://doi.org/10.5281/zenodo.12872948}.

The automatic screening consists of a pipeline with two Python packages: Pandas  \cite{OR10} and PyMuPDF\footnote{\url{https://pypi.org/project/PyMuPDF/}}. We first used Pandas to extract into a dataframe (i.e. table) all exported papers’ file locations and key BibTex or equivalent information including title, year, page number, DOI, and ISBN. Next, we used PyMuPDF to iterate through each PDF file and add to the dataframe multiple data fields: whether the file was successfully decoded\footnote{\url{https://pdfminersix.readthedocs.io/en/latest/faq.html}} for text extraction (if marked unsuccessful, the file was marked for manual screening), the occurrence count of each Regex keyword pattern listed below, and whether each keyword occurs after the section headings that fit into Regex patterns that represent variations of ``references'' and ``bibliography'' (referred to as ``non-target sections'' below). We then marked the files for exclusion by evaluating the eight criteria listed under ``Auto-excluded'' in \textbf{Table~\ref{table_a2_screen_steps}} against the information recorded in the dataframe. Each of the auto-exclusion results from Steps 1-4 and 7 in \textbf{Table~\ref{table_a2_screen_steps}} were manually checked, and those under Steps 5, 6, and 8 were spot-checked. These steps excluded 3277 out of the 4194 exported files.

Below are the regex patterns used for automatic keyword extraction and occurrence calculation:
    \begin{quote}
    PDF search keyword Regex list: ['absa', 'aspect$\backslash$W+base$\backslash$w*', 'aspect$\backslash$W+extrac$\backslash$w*', 'aspect$\backslash$W+term$\backslash$w*', 
    'aspect$\backslash$W+level$\backslash$w*', 
    'term$\backslash$W+level', 'sentiment$\backslash$W+analysis', 'opinion$\backslash$W+mining']
    \end{quote}

For the 914 files filtered through the auto-exclusion process, we manually screened them individually according to the inclusion and exclusion criteria. As shown in the second half of \textbf{Table~\ref{table_a2_screen_steps}}, this final screening step refined the review scope to 519 papers. 

\begin{table*}[ht!]
\centering
    \begin{threeparttable}[b]
  \caption{The automatic and manual screening processes}
  \label{table_a2_screen_steps}

\begin{tabular}{@{}lrr@{}}
    
\toprule
\textbf{Inclusion/Exclusion Types} & \multicolumn{1}{l}{\textbf{File Count}} & \multicolumn{1}{l}{\textbf{}} \\ \midrule
\textbf{Total Extracted Files}                                                & \textbf{} & \textbf{4191} \\ \midrule
\textbf{Auto-excluded}                                                        & \textbf{} & \textbf{3277} \\
\hspace{3mm} Step\_1 - Duplicate DOI across databases                                      & 19        &               \\
\hspace{3mm} Step\_2 - Duplicate Title across databases                                    & 11        &               \\
\hspace{3mm} Step\_3 - Survey/Not primary study papers                                     & 149       &               \\
\hspace{3mm} Step\_4 - Total keyword outside reference match = 0                           & 26        &               \\
\hspace{3mm} Step\_5 - keyword matched only to SA and/or OM outside reference \tnote{1}    & 2235      &               \\
\hspace{3mm} Step\_6 - No sentiment\textbackslash{}W+analysis in keyword outside reference & 243       &               \\
\hspace{3mm} Step\_7 - Total pages \textless 3                                             & 7         &               \\
\hspace{3mm} Step\_8 - Total keyword (except SA, OM) outside Reference \textless{} 5 \tnote{1,2}  & 587       &               \\
\textbf{Manually Excluded}                                                    & \textbf{} & \textbf{395}  \\ \midrule
\hspace{3mm} Type - Article withdrawn                                                      & 1         &               \\
\hspace{3mm} Type - Not published in English                                               & 1         &               \\
\hspace{3mm} Type - Dataset unclear                                                        & 1         &               \\
\hspace{3mm} Type - Low quality, unclear definition of ABSA                                & 1         &               \\
\hspace{3mm} Type - No original method                                                     & 2         &               \\
\hspace{3mm} Type - Duplicate of another included paper                                    & 3         &               \\
\hspace{3mm} Type - Sentence level SA                                                      & 4         &               \\
\hspace{3mm} Type - Not text-data focused                                                  & 6         &               \\
\hspace{3mm} Type - Not primary study                                                      & 15        &               \\
\hspace{3mm} Type - Lack of details on ABSA tasks                                          & 31        &               \\
\hspace{3mm} Type - Review (Not primary study)                                             & 43        &               \\
\hspace{3mm} Type - No ABSA task experiment details/results                                & 49        &               \\
\hspace{3mm} Type - Not ABSA focused; No ABSA task experiment details/results              & 238       &               \\ \midrule
\textbf{Final Included for Review}                                            & \textbf{} & \textbf{519}  \\ \bottomrule
\end{tabular}
\begin{tablenotes}
\item[1] SA, OM:  the Regex keyword patterns 'sentiment$\backslash$W+analysis', 'opinion$\backslash$W+mining' respectively. 
\item[2] The occurrence threshold 5 was chosen based on pilot file examination, which suggested that files with target keyword occurrence below this threshold tended to be non-ABSA-focused.
\end{tablenotes}
  \end{threeparttable}
\end{table*}

\subsection{Data extraction and synthesis} \label{sec_a3_domain_mapping}

In the final step of the SLR, we manually reviewed each of the 519 in-scope publications and recorded information according to a pre-designed data extraction form. The key information recorded includes each study’s research focus, research application domain (``research domain'' below),  ABSA subtasks involved, name or description of all the datasets directly used, model name (for machine-learning solutions), architecture, whether a certain approach or paradigm is present in the study (e.g. supervised learning, deep learning, end-to-end framework, ontology, rule-based, syntactic-components), and the specific approach used (e.g. attention mechanism, Naïve Bayes classifier) under the deep learning and traditional machine learning categories.  

After the data extraction, we performed data cleaning to identify and fix recording errors and inconsistencies, such as data entry typos and naming variations of the same dataset across studies. Then we created two mappings for the research and dataset domains described below. 

For each reviewed study, its research domain was defaulted to ``non-specific'' unless the study mentioned a specific application domain or use case as its motivation, in which case that domain description was recorded instead. 

The dataset domain was recorded and processed at the individual dataset level, as many reviewed studies used multiple datasets. We standardised the recorded dataset names, checked and verified the recorded dataset domain descriptions provided by the authors or the source web-pages, and then manually categorised each domain description into a domain category. For published/well-known datasets, we unified the recorded naming variations and checked the original datasets or their descriptions to verify the domain descriptions. For datasets created (e.g. web-crawled) by the authors of the reviewed studies, we named them following the ``[source] [domain] (original)'' format, e.g. ``Yelp restaurant review (original) '', or ``Twitter (original)'' if there was no distinct domain, and did not differentiate among the same-name variations. In all of the above cases, if a dataset was not created with a specific domain filter (e.g. general Twitter tweets), then it was classified as ``non-specific''.   

The recorded research and dataset domain descriptions were then manually grouped into 19 common domain categories. We tried to maintain consistency between the research and dataset domain categories. The following are two examples of possible mapping outcomes: 

\begin{enumerate}
    \item A study on a full ABSA solution without mentioning a specific application domain and using Yelp restaurant review and Amazon product review datasets would be assigned a research domain of ``non-specific'' and a dataset domain of ``product/service review''. 
    \item A study mentioning ``helping companies improve product design based on customer reviews'' as the motivation would have a research domain of ``product/service review'', and if they used a product review dataset and Twitter tweets crawled without filtering, the dataset domains would be ``product/service review'' and ``non-specific''.  
\end{enumerate} 

After applying the above-mentioned standardisation and mappings, we analysed the synthesised data quantitatively using the Pandas  \cite{OR10} library to obtain an overview of the reviewed studies and explore the answers to our RQs.


\newpage

\section{Additional Results}\label{appendix_C_results}

\setcounter{figure}{0}
\setcounter{table}{0}


\begin{figure*}[ht!]
  \centering
  \includegraphics[width=\linewidth]{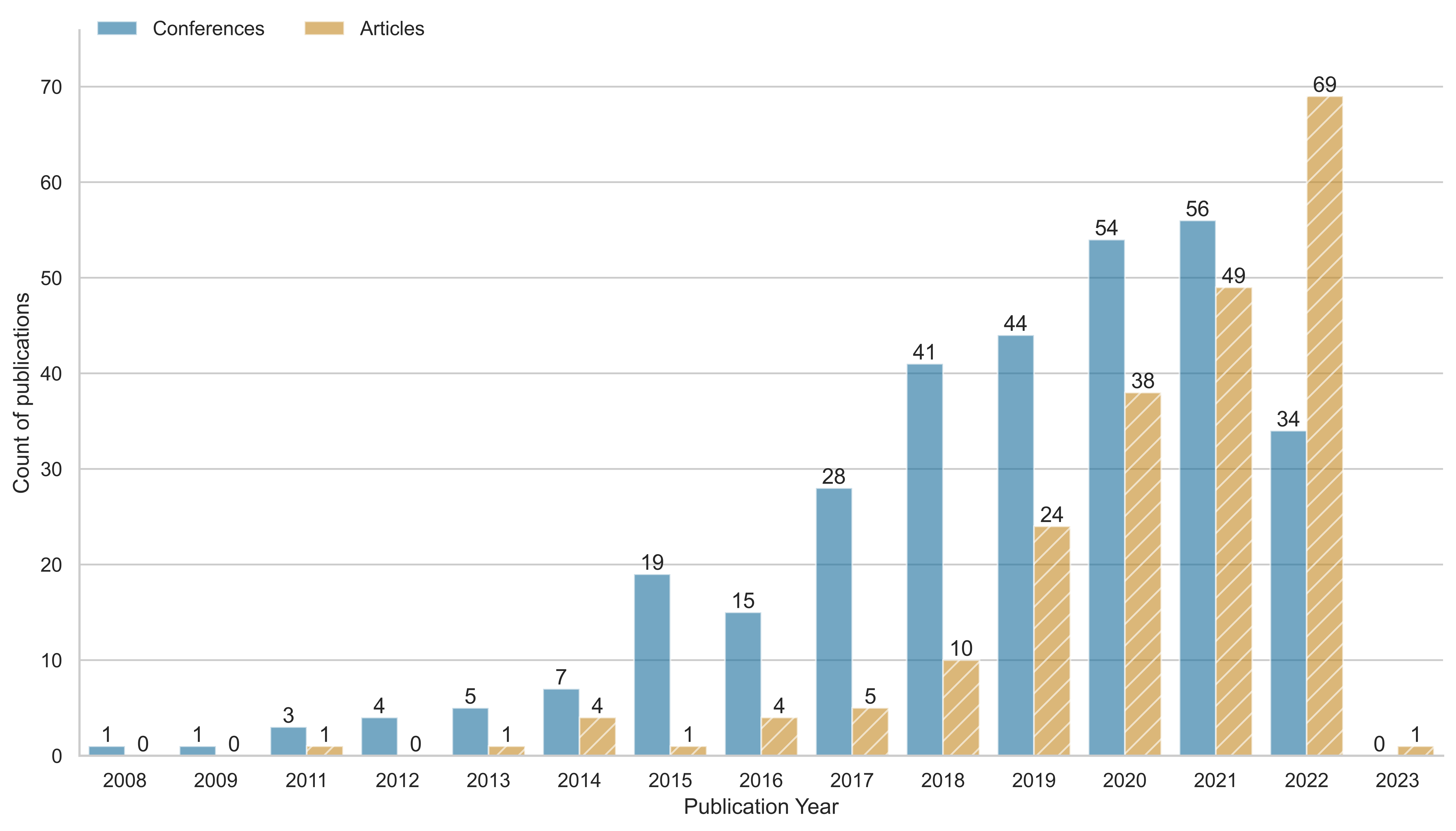}
  \caption{Number of Included Studies by Publication Year and Type (N=519) 
  \newline
  Note: Although our original search scope included journal articles, conference papers, newsletters, and magazine articles, the final 519 in-scope studies consist of only journal articles and conference papers. Conference papers noticeably outnumbered journal articles in all years until 2022, with the gap closing since 2016. We think this trend could be due to multiple factors, such as the fact that our search was conducted in late October 2022 when some conference publications were still not available; the publication lag for journal articles due to a longer processing period; and potentially a change in publication channels that is outside the scope of this review.
  }
  \label{fig_c1}
\end{figure*}

\begin{figure*}[!ht]
  \centering
  \includegraphics[width=\linewidth]{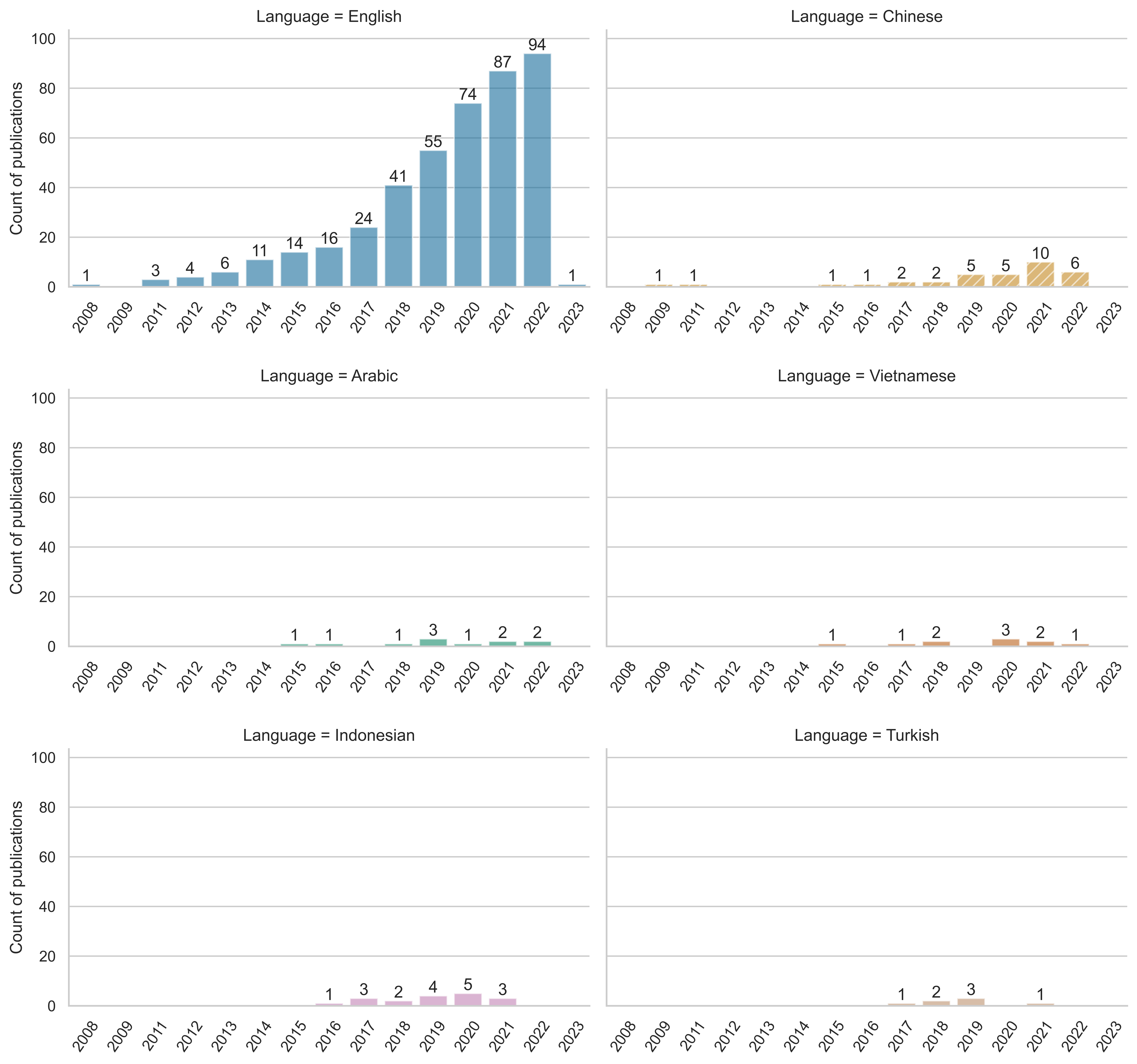}
  \caption{Number of Included Studies with the Top 5 Dataset Languages by Publication Year}
  \label{fig: fig_c2}
\end{figure*}

\begin{table}[ht!] 
\centering
    \caption{Number of Studies by ABSA Subtask Combinations (N=519)}
    \label{table6_tasks}
\begin{tabular}%
{>{\raggedright\arraybackslash}p{0.43\linewidth}%
 >{\raggedleft\arraybackslash}p{0.2\linewidth}%
 >{\raggedleft\arraybackslash}p{0.15\linewidth}}
\toprule

\textbf{Subtask Combination}                & \textbf{Count of Studies} & \textbf{\% of Studies} \\ \midrule
AE, ASC                                     & 168                       & 32.37\%                \\
ASC                                         & 160                       & 30.83\%                \\
AE                                          & 79                        & 15.22\%                \\
ACD, ASC                                    & 16                        & 3.08\%                 \\
ASTE                                        & 15                        & 2.89\%                 \\
AE, ACD, ASC                                & 13                        & 2.50\%                 \\
AE, OE, ASC                                 & 9                         & 1.73\%                 \\
AE, ACD                                     & 9                         & 1.73\%                 \\
AOPE                                        & 9                         & 1.73\%                 \\
ACD                                         & 7                         & 1.35\%                 \\
AE, OE                                      & 7                         & 1.35\%                 \\
AE, OE, ACD                                 & 4                         & 0.77\%                 \\
AE, OE, ACD, ASC                            & 4                         & 0.77\%                 \\
ASQE                                        & 2                         & 0.39\%                 \\
AOPE, ASC                                   & 2                         & 0.39\%                 \\
OE                                          & 1                         & 0.19\%                 \\
OTE (Opinion Target Extraction), ASC        & 1                         & 0.19\%                 \\
Aspect-based embedding, ACD, ASC            & 1                         & 0.19\%                 \\
ASC, OE                                     & 1                         & 0.19\%                 \\
Aspect and synthetic sample discrimination  & 1                         & 0.19\%                 \\
AE, OE, ASC, AOPE, ASTE                     & 1                         & 0.19\%                 \\
AOPE, ASC, ACD                              & 1                         & 0.19\%                 \\
AOPE, ACD                                   & 1                         & 0.19\%                 \\
AE, review-level SA                         & 1                         & 0.19\%                 \\
ACD, AE, ASC                                & 1                         & 0.19\%                 \\
AE, ASC, Aspect-based sentence segmentation & 1                         & 0.19\%                 \\
AE, ASC, Aspect-based embedding             & 1                         & 0.19\%                 \\
AE, ASC, ACD                                & 1                         & 0.19\%                 \\
AE, ACD, OE                                 & 1                         & 0.19\%                 \\
Data augmentation                           & 1                         & 0.19\%                 \\
\midrule
\textbf{TOTAL}                              & \textbf{519}              & \textbf{100.00\%}      \\
\bottomrule
\end{tabular}

\begin{tablenotes}
\item[1] This table corresponds to \textbf{Figure ~\ref{fig:fig7}} 
\end{tablenotes}

\end{table}

\begin{table} 
\centering
    \caption{Number of Studies by Individual ABSA Subtasks (N=805)}
    \label{table7_singletask}
\begin{tabular}%
{>{\raggedright\arraybackslash}p{0.43\linewidth}%
 >{\raggedleft\arraybackslash}p{0.2\linewidth}%
 >{\raggedleft\arraybackslash}p{0.15\linewidth}}
\toprule

\textbf{Individual Subtask}                & \textbf{Count of Studies} & \textbf{\% of Studies} \\ \midrule
ASC                                          & 381                       & 47.33\%                \\
AE                                           & 300                       & 37.27\%                \\
ACD                                          & 59                        & 7.33\%                 \\
OE                                           & 28                        & 3.48\%                 \\
ASTE                                         & 16                        & 1.99\%                 \\
AOPE                                         & 14                        & 1.74\%                 \\
ASQE                                         & 2                         & 0.25\%                 \\
Aspect and synthetic sample discrimination   & 1                         & 0.12\%                 \\
OTE (Opinion Target Extraction)              & 1                         & 0.12\%                 \\
Aspect-based sentence segmentation           & 1                         & 0.12\%                 \\
Data augmentation                            & 1                         & 0.12\%                 \\
Review-level SA                              & 1                         & 0.12\%                 \\
\midrule
\textbf{TOTAL}                               & \textbf{805}              & \textbf{100\%}         \\

\bottomrule
\end{tabular}

\begin{tablenotes}
\item[1] This table corresponds to \textbf{Figure ~\ref{fig:fig8}} 
\end{tablenotes}
\end{table}

\begin{table}[ht!] 
\centering
    \caption{Number of Studies by Deep Learning (DL) Approaches (N=313)}
    \label{table9_dl}
\begin{tabular}%
{>{\raggedright\arraybackslash}p{0.4\linewidth}%
 >{\raggedleft\arraybackslash}p{0.2\linewidth}%
 >{\raggedleft\arraybackslash}p{0.15\linewidth}}
\toprule

\textbf{DL Approach}        & \textbf{Count of Studies} & \textbf{\% of Studies} \\ \midrule
RNN, Attention              & 83                        & 26.52\%                \\
Attention                   & 60                        & 19.17\%                \\
RNN                         & 31                        & 9.90\%                 \\
RNN, CNN, Attention         & 25                        & 7.99\%                 \\
CNN                         & 20                        & 6.39\%                 \\
RNN, CNN                    & 19                        & 6.07\%                 \\
GNN/GCN, Attention          & 17                        & 5.43\%                 \\
CNN, Attention              & 17                        & 5.43\%                 \\
RNN, GNN/GCN, Attention     & 17                        & 5.43\%                 \\
Other (\textless{}5\% each) & 24                        & 7.67\%                 \\
\midrule
\textbf{TOTAL}              & \textbf{313}              & \textbf{100.00\%}      \\ 

\bottomrule
\end{tabular}

\begin{tablenotes}
\item[1] This table corresponds to \textbf{Figure ~\ref{fig:fig10}} 
\end{tablenotes}
\end{table}

\begin{table}[ht!] 
\centering
    \caption{Number of Studies by Traditional Machine Learning Approaches (N=283)}
    \label{table11_trad_ml}
\begin{tabular}%
{>{\raggedright\arraybackslash}p{0.4\linewidth}%
 >{\raggedleft\arraybackslash}p{0.2\linewidth}%
 >{\raggedleft\arraybackslash}p{0.15\linewidth}}
\toprule

\textbf{Traditional ML}           & \textbf{Count of Studies} & \textbf{\% of Studies} \\ \midrule
Support Vector Machine (SVM)      & 57                        & 20.14\%                \\
Conditional Random Field (CRF)    & 41                        & 14.49\%                \\
Latent Dirichlet allocation (LDA) & 36                        & 12.72\%                \\
Na\"ive Bayes (NB)                  & 32                        & 11.31\%                \\
Random Forest (RF)                & 24                        & 8.48\%                 \\
Decision Tree (DT)                & 18                        & 6.36\%                 \\
Logistic Regression (LR)          & 15                        & 5.30\%                 \\
K-Nearest Neighbors (KNN)         & 12                        & 4.24\%                 \\
Multinomial Na\"ive Bayes (MNB)     & 10                        & 3.53\%                 \\
K-means                           & 5                         & 1.77\%                 \\
Boosting                          & 5                         & 1.77\%                 \\
Other (N\textless{}5)             & 28                        & 9.89\%                 \\
\midrule
\textbf{TOTAL}                    & \textbf{283}              & \textbf{100.00\%}      \\

\bottomrule
\end{tabular}

\begin{tablenotes}
\item[1] This table corresponds to \textbf{Figure ~\ref{fig:fig11}} 
\end{tablenotes}
\end{table}

\clearpage 

\newpage

\centering
    
 
\begin{longtable}[ht!]{>{\raggedright\arraybackslash}p{0.61\linewidth}%
 >{\raggedleft\arraybackslash}p{0.12\linewidth}%
 >{\raggedleft\arraybackslash}p{0.1\linewidth}}
 
\caption{Number of Studies Per Dataset (N=1179, with 519 studies and 218 datasets)\label{tab:c5_dataset}}\\

\toprule
\textbf{Datasets}  & \textbf{Count of Studies} & \textbf{\% of Studies} \\ 
\midrule
\endfirsthead  

\caption{(continued)}\\

\toprule
\textbf{Datasets}  & \textbf{Count of Studies} & \textbf{\% of Studies} \\ 
\midrule

\endhead 

SemEval 2014 Restaurant                                        & 211                       & 17.90\%                \\
SemEval 2014 Laptop                                            & 189                       & 16.03\%                \\
SemEval 2016 Restaurant                                        & 118                       & 10.01\%                \\
SemEval 2015 Restaurant                                        & 106                       & 8.99\%                 \\
Twitter (Dong et al. 2014) \cite{dong-etal-2014-adaptive}                                   & 55                        & 4.66\%                 \\
Amazon customer review datasets (Hu \& Liu, 2004) \cite{HuLiu2004a}              & 33                        & 2.80\%                 \\
Amazon product review (original)                               & 23                        & 1.95\%                 \\
Product review (original)                                      & 22                        & 1.87\%                 \\
Twitter (original)                                             & 21                        & 1.78\%                 \\
SemEval 2015 Laptop                                            & 20                        & 1.70\%                 \\
Yelp Dataset Challenge Reviews                                 & 16                        & 1.36\%                 \\
SemEval 2016 Laptop                                            & 14                        & 1.19\%                 \\
TripAdvisor Hotel review (original)                            & 11                        & 0.93\%                 \\
Hotel review (original)                                        & 10                        & 0.85\%                 \\
TripAdvisor Restaurant review (original)                       & 9                         & 0.76\%                 \\
TripAdvisor Hotel review (Wang et al. 2010) \cite{wang-etal-2010-TripAdvisorHotel}                  & 9                         & 0.76\%                 \\
Movie review (original)                                        & 8                         & 0.68\%                 \\
MAMS Multi-Aspect Multi-Sentiment dataset (Jiang et al. 2019) \cite{OR18} & 8                         & 0.68\%                 \\
SemEval 2016 Hotel                                             & 7                         & 0.59\%                 \\
Twitter (Mitchell et al. 2013) \cite{mitchell-etal-2013-twitterdataset}   & 7                         & 0.59\%                 \\
Amazon product review (McAuley et al. 2015) \cite{McAuley-etal-2015-amazonDataset}   & 7                         & 0.59\%                 \\
VLSP 2018 Restaurant review                                    & 6                         & 0.51\%                 \\
Restaurant review (Ganu et al. 2009) \cite{Ganu2009}                         & 6                         & 0.51\%                 \\
Student feedback (original)                                    & 6                         & 0.51\%                 \\
VLSP 2018 Hotel review                                         & 6                         & 0.51\%                 \\
Chinese product review (Peng et al. 2018) \cite{peng-etal-2018-ChineseReview}                     & 5                         & 0.42\%                 \\
Stanford Twitter Sentiment / Sentiment140 (Go et al. 2009) \cite{Go2009TwitterSC}  & 5                         & 0.42\%                 \\
Yelp restaurant review (original)                              & 5                         & 0.42\%                 \\
SentiHood (Saeidi et al. 2016) \cite{saeidi-etal-2016-sentihood}                     & 4                         & 0.34\%                 \\
Game review (original)                                         & 3                         & 0.25\%                 \\
Sanders Twitter Corpus (STC) (Sanders, 2011) \cite{sander_twitter_stc}                  & 3                         & 0.25\%                 \\
TripAdvisor Tourist review (original)                          & 3                         & 0.25\%                 \\
Coursera course review (original)                              & 3                         & 0.25\%                 \\
Online drug review (original)                                  & 3                         & 0.25\%                 \\
Restaurant review (original)                                   & 3                         & 0.25\%                 \\
Chinese Restaurant review (original)                           & 3                         & 0.25\%                 \\
Financial Tweets and News Headlines dataset (FiQA 2018) \cite{fiqa_2018}      & 2                         & 0.17\%                 \\
SentiRuEval-2015                                               & 2                         & 0.17\%                 \\
Amazon laptop reviews (He \& McAuley, 2016) \cite{he-McAuley-2016-amazon}                   & 2                         & 0.17\%                 \\
Hindi ABSA dataset (Akhtar et al. 2016) \cite{akhtar-etal-2016-hindi-data}                      & 2                         & 0.17\%                 \\
Hotel review (Kaggle)                                          & 2                         & 0.17\%                 \\
Service review (Toprak etc. 2010) \cite{toprak-etc-2010-serviceReview}           & 2                         & 0.17\%                 \\
Indonesian Product review (original)                           & 2                         & 0.17\%                 \\
Indonesian Tweets - Political topic (original)                 & 2                         & 0.17\%                 \\
Steam game review (original)                                   & 2                         & 0.17\%                 \\
Stanford Sentiment Treebank review data (Socher, 2013) \cite{socher-etal-2013-recursive}        & 2                         & 0.17\%                 \\
Kaggle movie review                                            & 2                         & 0.17\%                 \\
Zomato restaurant review (original)                            & 2                         & 0.17\%                 \\
Turkish Product review (original)                              & 2                         & 0.17\%                 \\
SemEval 2015 Hotel                                             & 2                         & 0.17\%                 \\
Weibo comments (original)                                      & 2                         & 0.17\%                 \\
Amazon 50-domain Product review (Chen \& Liu, 2014) \cite{chenLiu-2014-50domaindata}            & 2                         & 0.17\%                 \\
Bangla ABSA Cricket, Restaurant dataset (Rahman \& Dey, 2018) \cite{rahman-dey-2018-bangleDataset}   & 2                         & 0.17\%                 \\
YouTube comments (original)                                    & 2                         & 0.17\%                 \\
Beer review (McAuley et al. 2012) \cite{McAuley-etal-2012-beerreview}                      & 2                         & 0.17\%                 \\
CCF BDCI 2018 Chinese auto review dataset                      & 2                         & 0.17\%                 \\
ReLi Portuguese book reviews (Freitas, et al. 2012) \cite{freitas-etal-2014-reli}          & 2                         & 0.17\%                 \\
Twitter product review (original)                              & 2                         & 0.17\%                 \\
SemEval 2016 Tweets                                            & 2                         & 0.17\%                 \\
Product review dataset (Liu et al. 2015) \cite{liu-etal-2015-dataset}                     & 2                         & 0.17\%                 \\
Product review dataset (Cruz-Garcia et al. 2014) \cite{cruz-etal-2014-implicit}             & 2                         & 0.17\%                 \\
Chinese Product review (original)                              & 2                         & 0.17\%                 \\
ICLR Open Reviews dataset                                      & 2                         & 0.17\%                 \\
Amazon Electronics review data (Jo et al. 2011) \cite{jo-etal-2011-amazondata}                     & 2                         & 0.17\%                 \\
Amazon product review (Wang et al. 2011) \cite{wang-etal-2011-amazon-dataset}                     & 2                         & 0.17\%                 \\
SemEval 2015 (unspecified)                                     & 2                         & 0.17\%                 \\
Product \& service review (Kaggle)                             & 2                         & 0.17\%                 \\
Othe datasets (N=1 each)                                     & 149                       & 12.64\%                 \\

\midrule
\textbf{TOTAL}                                                 & \textbf{1179}             & \textbf{100.00\%}      \\

\bottomrule
\end{longtable}


\clearpage
\bibliographystyle{unsrt} 
\bibliography{SLR_ABSA_text.bbl}

\end{document}